\documentclass[10pt,journal,compsoc]{IEEEtran}
\usepackage{ifpdf}

\ifCLASSOPTIONcompsoc
  \usepackage[nocompress]{cite}
\else
  \usepackage{cite}
\fi
\ifCLASSINFOpdf
  \usepackage[pdftex]{graphicx}
  \graphicspath{{../pdf/}{../jpeg/}}
  \DeclareGraphicsExtensions{.pdf,.jpeg,.png}
\else
  \usepackage[dvips]{graphicx}
  \graphicspath{{../eps/}}
  \DeclareGraphicsExtensions{.eps}
\fi
\usepackage{amsmath}
\usepackage{algorithmic}

\usepackage{array}

\ifCLASSOPTIONcompsoc
 \usepackage[caption=false,font=footnotesize,labelfont=sf,textfont=sf]{subfig}
\else
 \usepackage[caption=false,font=footnotesize]{subfig}
\fi

\usepackage{stfloats}
\ifCLASSOPTIONcaptionsoff
 \usepackage[nomarkers]{endfloat}
\let\MYoriglatexcaption\caption
\renewcommand{\caption}[2][\relax]{\MYoriglatexcaption[#2]{#2}}
\fi
\usepackage{url}

\usepackage{float}
\usepackage{booktabs}
\usepackage{amssymb}
\usepackage{multirow}
\usepackage{multicol}
\usepackage{bm}
\usepackage{colortbl}
\usepackage{pifont}
\usepackage{enumitem}
\setlist[itemize]{leftmargin=0.5cm}

\makeatletter
\newcommand{\thickhline}{%
    \noalign {\ifnum 0=`}\fi \hrule height 1.2pt
    \futurelet \reserved@a \@xhline
}

\makeatletter
\long\def\@IEEEtitleabstractindextextbox#1{\parbox{0.922\textwidth}{#1}}
\makeatother

\definecolor{mygray}{gray}{.9}
\newcommand\ourmethod{DAT}
\newcommand\omlong{Deformable Attention Transformer}
\usepackage[pagebackref,breaklinks,colorlinks]{hyperref}

\usepackage[capitalize]{cleveref}
\crefname{section}{Sec.}{Secs.}
\Crefname{section}{Section}{Sections}
\Crefname{table}{Table}{Tables}
\crefname{table}{Tab.}{Tabs.}

\begin{document}
%
\title{DAT++: Spatially Dynamic Vision Transformer with Deformable Attention}
%
%
%
%

\author{Zhuofan~Xia, Xuran~Pan,
        Shiji~Song,
        Li~Erran~Li,
        and~Gao~Huang
\IEEEcompsocitemizethanks{\IEEEcompsocthanksitem Zhuofan Xia, Xuran Pan, Shiji Song and Gao Huang are with the Department of Automation, BNRist, Tsinghua University, Beijing 100084, China. Gao Huang is also with Beijing Academy of Artificial Intelligence (BAAI). Corresponding author: Gao Huang.\protect\\
E-mail: xzf23@mails.tsinghua.edu.cn.
\IEEEcompsocthanksitem Li Erran Li is with AWS AI, Amazon, Santa Clara, CA 95054.
\IEEEcompsocthanksitem Code is publicly available at \url{https://github.com/LeapLabTHU/DAT}. 
}
}

\IEEEtitleabstractindextext{%
\begin{abstract}
Transformers have shown superior performance on various vision tasks. Their large receptive field endows Transformer models with higher representation power than their CNN counterparts. Nevertheless, simply enlarging the receptive field also raises several concerns. On the one hand, using dense attention in ViT leads to excessive memory and computational cost, and features can be influenced by irrelevant parts that are beyond the region of interests. On the other hand, the handcrafted attention adopted in PVT or Swin Transformer is data agnostic and may limit the ability to model long-range relations. To solve this dilemma, we propose a novel deformable multi-head attention module, where the positions of key and value pairs in self-attention are adaptively allocated in a data-dependent way. This flexible scheme enables the proposed deformable attention to dynamically focus on relevant regions while maintains the representation power of global attention. On this basis, we present \textbf{\omlong{} (\ourmethod{})}, a general vision backbone efficient and effective for visual recognition. We further build an enhanced version \ourmethod{}++. Extensive experiments show that our \ourmethod{}++ achieves state-of-the-art results on various visual recognition benchmarks, with 85.9\% ImageNet accuracy, 54.5 and 47.0 MS-COCO instance segmentation mAP, and 51.5 ADE20K semantic segmentation mIoU.
\end{abstract}

\begin{IEEEkeywords}
Vision Transformer, deformable attention, dynamic neural networks.
\end{IEEEkeywords}
}

\maketitle

\IEEEdisplaynontitleabstractindextext

%
\IEEEpeerreviewmaketitle

\IEEEraisesectionheading{\section{Introduction}\label{sec:introduction}}

\IEEEPARstart{T}{ransformer}~\cite{attention} is originally introduced to solve natural language processing tasks. It has recently shown great potential in the field of computer vision~\cite{ViT16x16, Swin, PVT}. The pioneering work, Vision Transformer~\cite{ViT16x16} (ViT), stacks multiple Transformer blocks to process non-overlapping image patch (\textit{i.e.} visual token) sequences, leading to a family of convolution-free models for visual recognition. Compared to their CNN counterparts~\cite{resnet,densenet,regnet,mobilev1,mobilev2,mobilev3,condensev1,condensev2}, ViTs have larger receptive fields and excel at modeling long-range dependencies, which are proven to achieve superior performance in the regime of a large amount of training data and model parameters. However, superfluous attention in visual recognition is a double-edged sword and has multiple drawbacks. Specifically, the excessive number of keys to attend per query patch yields high computational cost and slow convergence, and also increases the risk of overfitting.

\begin{figure}
    \begin{center}
    \vspace{-1cm}
    \includegraphics[width=0.97\linewidth]{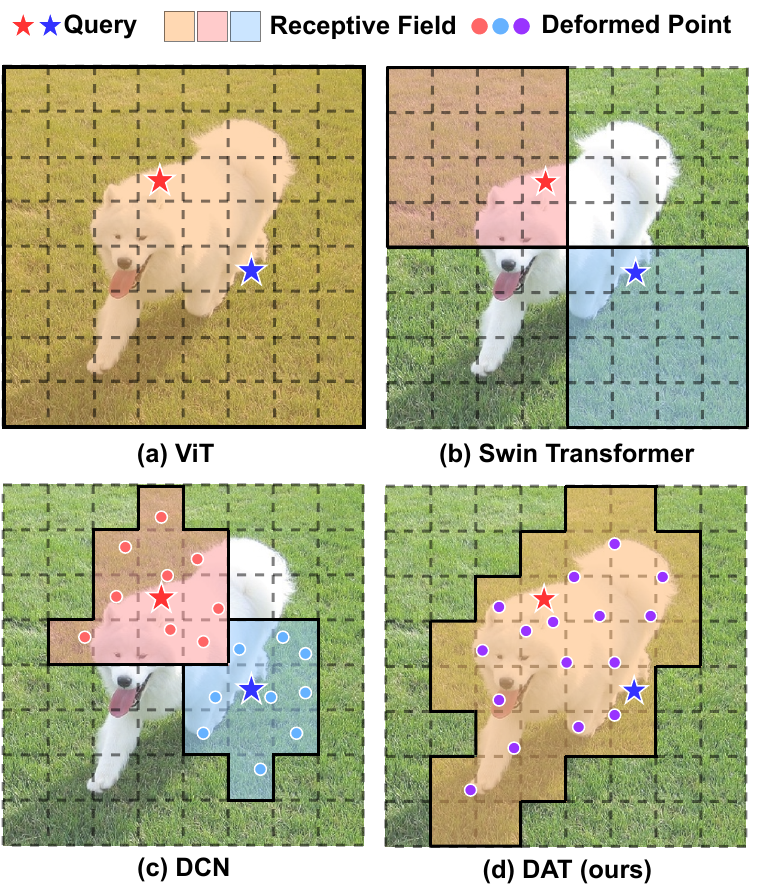}
    \vspace{-0.2cm}
    \caption{Comparison of DAT with other Vision Transformers and DCN~\cite{DCNv1}. The red star and the blue star denote the different queries, and masks with solid line boundaries depict the regions to which the queries attend. In a \textbf{data-agnostic} way: (a) ViT~\cite{ViT16x16} adopts full global attention for all queries. (b) Swin Transformer~\cite{Swin} uses partitioned window attention. In a \textbf{data-dependent} way: (c) DCN learns different deformed points for each query. (d) \ourmethod{} learns shared deformed key locations for all queries.}
    \label{fig:fig1}
    \end{center}
\end{figure}

In order to avoid excessive attention computation, existing works~\cite{PVT,Swin,vil,cswin,focal,rgvit,hrformer,mvit,mvitv2,PVTv2,tcformer} have leveraged carefully designed efficient sparse attention patterns to reduce computational complexity. As two representative approaches among them, Swin Transformer~\cite{Swin} adopts window-based local attention to restrict attention in local windows and shifts windows layer-wise to interact between windows, while Pyramid Vision Transformer (PVT)~\cite{PVT} spatially downsamples the key and value feature maps to save computation by attending queries to a coarsened set of keys. Although effective, hand-crafted attention patterns are data-agnostic and may not be optimal for the various images. It is likely that relevant keys/values are dropped while less important ones are still kept. 

Ideally, one would expect that the candidate key/value set for a given query is flexible and has the ability to adapt to each individual input, such that the issues in hand-crafted sparse attention patterns can be alleviated. In fact, in the literature on CNNs, learning deformable receptive fields for the convolution filters has been shown effective in selectively attending to more informative regions on a data-dependent basis, as depicted in Figure~\ref{fig:fig1}(c). The most notable work, Deformable Convolution Networks~\cite{DCNv1}, has yielded impressive results on many challenging vision tasks. This motivates us to explore a deformable attention mechanism in Vision Transformers. However, a naive implementation of this idea leads to an unreasonably high memory/computation complexity: the overhead introduced by the deformable offsets is quadratic \textit{w.r.t} the number of visual tokens, prohibiting the applications on high-resolution feature maps by huge memory footprints. As a consequence, although some recent works~\cite{DeformableDETR, dpt, psvit} have investigated the idea of deformable computation in Transformers, none of them has treated it as a fundamental building block to construct a powerful backbone network like the DCN, due to the high computational cost. Among these works, the deformable mechanism is either adopted in the detection head~\cite{DeformableDETR}, or used as a pre-processing module to sample patches for the subsequent backbone network~\cite{dpt,psvit}. 

In this paper, we present a simple and efficient deformable multi-head attention module (DMHA), equipped with which a powerful pyramid vision backbone, named \textbf{\omlong{} (\ourmethod{})}, is constructed for various visual recognition tasks, including image classification, object detection and instance segmentation, and semantic segmentation, \textit{etc}. Different from DCN, which learns different offsets for different pixels in the whole feature map, we propose to learn a few groups of \textbf{query agnostic} offsets, shifting keys and values to capture important regions (as illustrated in Figure~\ref{fig:fig1}(d)), based on the observation in~\cite{gcnet,deepvit,FCViT} that global attention usually results in almost the same attention patterns for different queries. This design introduces a data-dependent sparse attention pattern to Vision Transformer backbones while maintaining linear space complexity. Specifically, for each attention module, the reference points are first generated as uniform grids, which are the same across the input data. Then, the offset generation network takes the query features as input and predicts the corresponding offsets for all reference points. In this way, the candidate keys/values are shifted towards important regions, thus augmenting the original multi-head self-attention with higher flexibility and efficiency to capture more informative features. To achieve stronger performance, we adopted several advanced convolutional enhancements to build \textbf{\ourmethod{}++}, an improved version of \ourmethod{}. 

Extensive experiments on various visual recognition tasks demonstrate that our \ourmethod{}++ outperforms or achieves competitive performance compared to other state-of-the-art ViTs and CNNs. Specifically, the largest model \ourmethod{}-B++ achieves 85.9\% Top-1 accuracy on ImageNet~\cite{in1k} image classification, 54.5 bbox mAP and 47.0 mask mAP on MS-COCO~\cite{coco} instance segmentation, and 51.5 mIoU on ADE20K~\cite{ade20k} semantic segmentation, exhibiting the superiority of \ourmethod{}++. 


A preliminary version of this work was published in~\cite{Xia_2022_CVPR}. In this paper, we extend our conference version in the following aspects: 

\begin{itemize}
    \item We simplify the designs and remove several hyper-parameters in the original DMHA of \ourmethod{}, leading to a more concise and efficient implementation. We remove the constraint range factor on the learned offsets, extend the DMHA module to all stages of the model, and unify the number of deformed keys across stages.
    \item We propose an enhanced version of \ourmethod{} by incorporating overlapped patch embedding and downsampling, injecting depth-wise convolution into MLP blocks, and adopting local perception units, which enhance local features and positional information, and benefit the learning of the proposed deformable attention. These improvements lead to \ourmethod{}++, a spatial dynamic vision foundation model, which achieves state-of-the-art performance in many challenging vision tasks.
    \item We provide more analytical experiments and visualization to further study the effectiveness of \ourmethod{} and \ourmethod{}++. The ablation experiments are extended with the results of object detection and semantic segmentation. In addition to the study of the core components in DMHA, we sketch a roadmap from \ourmethod{} to \ourmethod{}++ as shown in Figure~\ref{fig:roadmap}, to analyze the effect of each modification in detail. We also show the versatility of the proposed DMHA to various local modeling operators, including different types of local attention and convolution. To distinguish our deformable attention from the one in Deformable DETR, we provide comprehensive comparisons both theoretically and empirically. In addition to quantitative analyses, more exquisite visualization are presented to verify the effectiveness of \ourmethod{}++.
\end{itemize}

\section{Related Works}\label{sec:related_works}

\subsection{Vision Transformers}\label{sec:r_vit}

Since the introduction of ViT~\cite{ViT16x16}, improvements~\cite{PVT,Swin,vil,cswin,mvit,hrformer,hrvit,vitae,scalablevit,maxvit,davit,crossformerpp,svit,biformer} have focused on learning multiscale features with hierarchical architecture and efficient spatial attention mechanisms. These attention mechanisms include windowed attention~\cite{Swin, cswin, Swinv2, scalablevit}, pooling attention~\cite{mvit, mvitv2}, global or region tokens~\cite{vil, rgvit, vparser, perceiver, davit, msg}, multiscale feature inductive biases~\cite{focal, focalnet, pervit, crossformerpp, vitae, vitaev2}, and high-resolution architectures~\cite{hrformer,hrvit} similar to HRNet~\cite{hrnet}. In addition to hierarchical architecture, there have also been many attempts to enhance isotropic ViT in many ways, \textit{e.g.}, introducing knowledge distillation~\cite{deit, cait, deepvit, tinyvit ,lvvit}, improving data augmentations~\cite{3things,deit3,howtotrain}, exploring self-supervised learning in Contrastive Learning~\cite{mocov3,dino,dinov2,flip} and Mask Image Modeling~\cite{simmim,mae,beit,ibot,maskfeat,m3i}, and scaling ViTs to larger models~\cite{eva,ijepa,vit22b,modelsoup}.

In addition to the development of model architectures or learning targets, it is a growing trend that incorporating dynamic neural networks~\cite{survey} into ViT can better exploit the spatial structure of input images and significantly improve ViT efficiency. Among these algorithms, dynamic length of visual tokens~\cite{dvt,dynamicvit,evovit,adavit,avit,evit,tome} aims to determine the vital patch tokens in a data-dependent way and reduce the total length by pruning or merging tokens. Apart from dynamic length, another line of research investigates the dynamic mechanisms in attention modeling, such as dividing image space into trees~\cite{quadtree,deptree}, clustering or predicting super tokens~\cite{tcformer,dge,svit,clustr}, sparse bi-level efficient attention~\cite{biformer}, and adapting attention window shapes~\cite{vsa,quadrangle}. In this work, we propose a novel dynamic attention mechanism for visual recognition with deformable attention by learning offsets to deform keys and values, encouraging the deformable attention to focus the important regions in the image.

\subsection{Convolution-based Models}\label{sec:r_cnn}

Recently, convolution-based approaches have been introduced into Vision Transformer models~\cite{coat,coatnet,cmt,san,acmix,slide}. 
ConViT~\cite{convit} develops a convolutional inductive bias alongside the attention branch. CvT~\cite{cvt} adopts overlapped convolutions in patch embedding and projection layers in attention, and CPE~\cite{cpvt} proposes a convolutional position embedding. Based on these early discoveries, more comprehensive works, including CoaT~\cite{coat}, CoAtNet~\cite{coatnet}, and CMT~\cite{cmt}, incorporate more convolution modules and achieve better results. Some analyses~\cite{early,connect} also point out the relation of convolutions and local attention, inspiring recent efficient designs of local visual attention, such as QnA-ViT~\cite{qna}, NAT~\cite{nat,dinat}, and Slide Transformer~\cite{slide}. In this work, we apply some convolution-based enhancements on top of \ourmethod{} to build an improved version, \ourmethod{}++. 

In addition to the convolution-enhanced ViT, recent progress of ViT has inspired several novel CNNs with block structures similar to ViT and larger convolution kernel sizes. ConvNeXt~\cite{convnext} adjusts the conventional CNN structure in both micro and macro designs, building a strong CNN model comparable to Swin Transformer. RepLKNet~\cite{replknet} and SLaK~\cite{slak} take a step towards larger kernel sizes with structural reparameterization, while HorNet~\cite{hornet} presents another method to expand the receptive field by higer-order convolutions with filters in the global frequency domain~\cite{gf}. InternImage~\cite{DCNv3} proposes a strong foundation model based on improved deformable convolution, whose largest variant reaches an unprecedented level on MS-COCO object detection. Extensive experiments show \ourmethod{}++ can achieve better or comparable performance to the state-of-the-art ViT/CNNs on various tasks.

\subsection{Deformable CNN and Attention}\label{sec:r_dcn}

Deformable convolution~\cite{DCNv1, DCNv2, DCNv3} is a powerful dynamic convolution mechanism to attend to flexible spatial locations conditioned on the structure of the input image. Recently, it has been applied to Vision Transformers~\cite{DeformableDETR,dpt,psvit,sparse}. Deformable DETR~\cite{DeformableDETR} improves the convergence of DETR~\cite{detr} by attending each query to a small number of deformed keys on the top of a CNN backbone. DPT~\cite{dpt}, PS-ViT~\cite{psvit}, and SparseFormer~\cite{sparse} build deformable modules to refine visual tokens and pass these tokens to the following isotropic ViT. Specifically, DPT proposes a deformable patch embedding module instead of regular downsampling to refine the patches to the next model stages dynamically. PS-ViT introduces a spatial sampling module before a ViT backbone to sample visual tokens on the crucial parts of the image. SparseFormer improves PS-ViT by first selecting ROIs in the image and then sampling important points in the ROIs to decode latent tokens, which involves a coarse-to-fine manner to promote efficiency. However, none of these works incorporates deformable attention into vision backbones. On the contrary, our deformable attention takes a powerful yet simple design to learn a set of global keys shared among visual tokens and can be adopted as a general backbone for various vision tasks. Our method can also be viewed as a spatial adaptive mechanism, which has been proven effective in various works~\cite{gfnet,sarnet,arc,dynper}.

\begin{figure*}[t]
    \begin{center}
    \includegraphics[width=\linewidth]{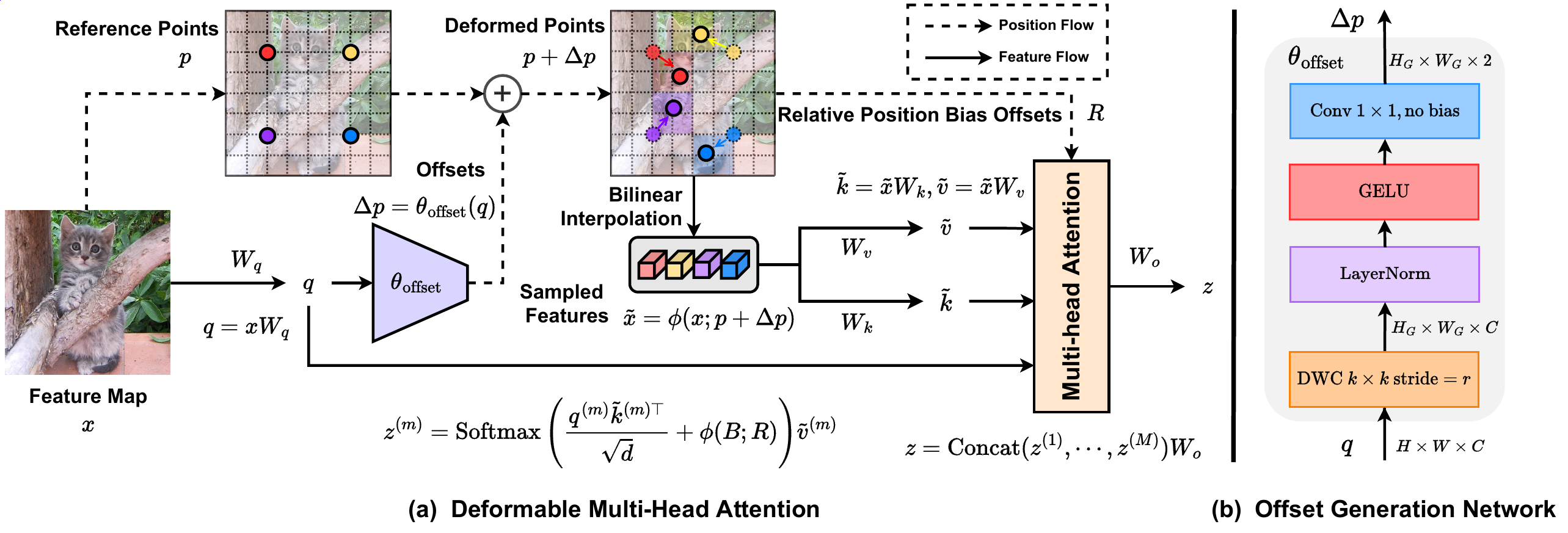}
    \vspace{-0.8cm}
    \caption{An illustration of proposed Deformable Attention mechanism. (a) presents the information flow of deformable multi-head attention (DMHA). On the left, a group of reference points is placed uniformly on the feature map, whose offsets are learned from the queries by the offset generation network. The features of important regions are sampled according to the deformed points with bilinear interpolation, which is depicted in the right part. The deformed keys and values are projected by the sampled features and then participate in computing attention. Relative position biases are also computed between the deformed keys and the query grid, enhancing multi-head attention with positional information. We show only 4 reference points for a clear display. (b) reveals the detailed structure of the offset generation network, marked with sizes of feature maps. The query feature map is first transformed by a depth-wise convolution to incorporate local information and then downsampled to the spatial size of the reference points. Another $1\!\times{}\!1$ convolution following normalization and activation transforms the feature map to the offset values.}
    \label{fig:fig2}
    \vspace{-0.5cm}
    \end{center}
\end{figure*}
\section{Deformable Attention Transformer}\label{sec:method}

\subsection{Preliminaries}\label{sec:m_pre}

We first revisit the attention mechanism in popular Vision Transformers (ViT)~\cite{ViT16x16}, as shown in Figure~\ref{fig:fig1}(a). Typically, patch tokens in the shape of $H\!\times{}\!W\!\times{}\!C$ are flattened to a sequence $x\in\mathbb{R}^{N\!\times{}\!C}$, whose length is $N\!=\!HW$. Taking the patch sequence as input, a vanilla multi-head self-attention (MHSA) block with $M$ attention heads is formulated as 
\begin{align}
    &q=xW_q,\hspace{15pt}k=xW_k,\hspace{15pt}v=xW_v, \label{eq:proj} \\
    &z^{(m)}=\sigma(q^{(m)}k^{(m)\top}/\sqrt{d})v^{(m)},\hspace{6pt}m=1,\ldots,M, \label{eq:attn} \\
    &z=\text{Concat}\left(z^{(1)},\ldots,z^{(M)}\right)W_o, \label{eq:MHSA}
\end{align}
where $\sigma(\cdot)$ denotes the softmax function, and $d\!=\!C/M$ is the dimension of each head. We define $z^{(m)}$ as the aggregated attention values at the $m$-th head and $q^{(m)}, k^{(m)}, v^{(m)}\in\mathbb{R}^{N\times{}d}$ to be query, key, and value at the $m$-th attention head, respectively. The query, key, value, and output projections are parameterized by $W_q,W_k,W_v,W_o\in\mathbb{R}^{C\times{}C}$ with the bias terms omitted. To build a Transformer block, an MLP block with two linear transformations and a GELU activation is usually adopted to provide nonlinearity.

With normalization layers and identity shortcuts, the $l$-th Vision Transformer block is formulated as
\begin{align}
    z_l'&=\text{MHSA}\left(\text{LN}(z_{l-1})\right)+z_{l-1}, \label{eq:att_layer} \\
    z_l&=\text{MLP}\left(\text{LN}(z_l')\right)+z_l', \label{eq:mlp_layer}
\end{align}
where LN is Layer Normalization~\cite{LN}. Since each token in ViT aggregates features from the entire set of tokens on the feature map, the vanilla MHSA enjoys a global receptive field and excels at modeling long-range spatial dependencies. However, global attention yields a high computational cost of $2(HW)^2C$ in terms of time complexity, restricting the computation to small feature maps with a stride of 16. Vanilla ViT also increases the risk of overfitting, requiring either larger-scale pretraining data~\cite{ViT16x16,modelsoup} or more sophisticated data augmentations~\cite{deit,howtotrain}.

Existing popular hierarchical Vision Transformers, notably PVT~\cite{PVT} and Swin Transformer~\cite{Swin}, try to address the challenge of excessive attention. PVT modifies global attention by downsampling the keys and values into smaller spatial sizes, whereas Swin Transformer enforces the attention into local windows, as depicted in Figure~\ref{fig:fig1}(b). The downsample of the former results in severe information loss, and the shift-window attention of the latter leads to a much slower growth of receptive fields, which limits the potential of modeling objects lying across multiple windows. 

\subsection{Deformable Attention}\label{sec:m_def}

\subsubsection{Deformable Convolution and Deformable Attention}\label{sec:m_def_intro}

The demand for data-dependent sparse attention is hence raised to model spatial relevant features of the tokens more flexibly. A similar methodology is initially proposed by Deformable Convolution Networks~\cite{DCNv1} in the field of convolutions, which learns the locations of the kernel weights in a data-dependent way. However, migrating the deformable convolution from CNNs to deformable attention Vision Transformers is a nontrivial problem.

In DCN, each element on the feature map learns its offsets individually, as shown in Figure~\ref{fig:fig1}(c), of which a $3\!\times{}\!3$ deformable convolution on an $H\!\times{}\!W\!\times{}\!C$ feature map has a space complexity of $9HWC$. If the same mechanism is applied directly to attention, the space complexity to store queries, keys, and values would drastically rise to $3N_\text{q}N_\text{k}C$, where $N_\text{q}, N_\text{k}$ are the numbers of queries and keys which have the same scale as the feature map size $HW$, bringing approximately a quadratic complexity of $3(HW)^2C$.

Although Deformable DETR~\cite{DeformableDETR} has managed to reduce the overhead by setting a lower number of keys with $N_\text{k}\!=\!4$ at each scale and producing attention weights from linear projections of the queries, it is inferior to attend to such few keys with linear attention in a backbone network due to the unacceptable loss of information. The detailed comparison and discussion are included in Sec.~\ref{sec:ddetr}.

In the meantime, the observations in many recent pieces of research~\cite{deepvit, gcnet, FCViT} reveal that different queries among the visual tokens possess similar attention maps in visual attention models, particularly in the deep layers. Furthermore, only a few keys dominate the attention scores in the above attention maps, indicating that dense attention between queries and keys may be dispensable. Therefore, a simple solution where each query shares the same deformed locations of sparse keys could achieve an efficient trade-off.

\begin{figure*}
    \begin{center}
    \includegraphics[width=\linewidth]{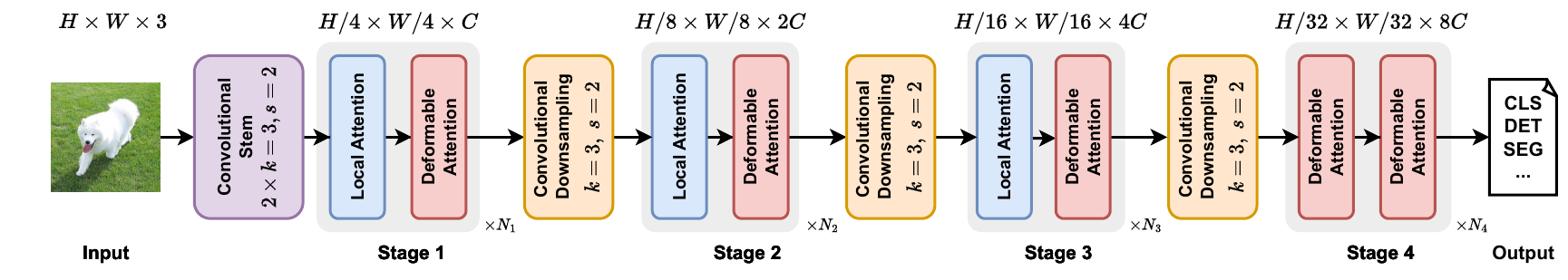}
    \vspace{-0.8cm}
    \caption{An illustration of the model architecture of \omlong{}++. $N_1$ to $N_4$ are the numbers of stacked successive local and deformable attention blocks. $k$ and $s$ denote the kernel size and stride of the convolution layer in patch embeddings. The stem consists of two consecutive layers of convolution-normalization-activation, whereas a normalization layer usually follows the convolution in downsampling blocks.}
    \label{fig:fig_model_arch}
    \end{center}
    \vspace{-0.4cm}
\end{figure*}

\subsubsection{Deformable Multi-Head Attention (DMHA)}\label{sec:m_def_attn}

We herein propose a novel \textbf{deformable multi-head attention (DMHA)} mechanism to effectively model the relations among visual tokens under the guidance of learned pivotal regions on the feature maps. These regions are determined by multiple groups of deformed sampling points whose locations are learned from the queries by an offset generation network. 
For each input image, the deformed points are individually learned to capture the important parts, thus encouraging a data-dependent sparse attention mechanism.
By this means, the challenge of excessive attention is tackled along with a relatively low space complexity in that no extra memory is allocated to store keys and values.

The design of DMHA is illustrated in Figure \ref{fig:fig2}(a). From the learned deformed points, the features in the important regions are sampled through bilinear interpolations and then fed to the key and value projections to obtain the deformed keys and values. Finally, the multi-head attention from queries to the sampled keys is applied to aggregate features from the deformed values. Additionally, the locations of deformed points provide a more powerful relative position bias to facilitate the learning of DMHA. These components are discussed in detail below.

\noindent
\textbf{Reference points.} First, a uniform grid $p\in\mathbb{R}^{H_G\!\times{}\!W_G\!\times{}\!2}$ is generated as the reference points on the input feature map $x\in\mathbb{R}^{H\!\times{}\!W\!\times{}\!C}$.
The grid is downsampled from the input feature map by a factor $r$, in the size of $H_G\!=\!H/r,W_G\!=\!W/r$. The reference points are linearly spaced at 2D lattice coordinates in
\begin{equation}
\{(p_x,p_y)|(0,0),\ldots,(W_G\!-\!1,H_G\!-\!1)\}.\label{eq:ref_pts}
\end{equation} 
These locations are then scaled into the range $[-1,+1]$ according to the reference shape $H_G\!\times{}\!W_G$ for numerical stability. At this scale, $(-1,-1)$ indicates the top left corner, and $(+1,+1)$ indicates the bottom right corner. 

\noindent
\textbf{Offsets learning.} To learn the spatial dynamic keys and values conditioned on the queries, we leverage a lightweight offset generation network $\theta_\text{offset}(\cdot)$ to predict the offsets from query tokens $q$: 
\begin{equation}
\Delta{}p=\theta_\text{offset}(q),\label{eq:delta_p_ogn}
\end{equation}
whose shape is identical to the reference points in $\Delta{}p\in\mathbb{R}^{H_G\!\times{}\!W_G\!\times{}\!2}$. The offsets are added to the positions of the reference points to determine the sampling locations of the deformed points. To avoid sampling outside the feature maps, we clip the sampling locations by $-1$ and $+1$ to ensure that all the pivotal sampling points lie in the region of the image. Different from the conference version~\cite{Xia_2022_CVPR}, the offset range factor is removed for a concise implementation.

\noindent
\textbf{Bilinear sampling.} To sample features from the important regions, a sampling function $\phi(\cdot;\cdot)$ with bilinear interpolation is adopted in support of the differentiability:
\begin{equation}
    \phi\left(z;(p_x,p_y)\right)\!=\!\sum_{(r_x,r_y)}\!g(p_x,r_x)g(p_y,r_y)z[r_y,r_x,:], \label{eq:bilinear}
\end{equation}
where $g(a,b)=\max(0,1-|a-b|)$ is the bilinear sampling weight function and $(r_x,r_y)$ indexes all the locations on $z\in\mathbb{R}^{H\!\times{}\!W\!\times{}\!C}$. As $g$ would be a nonzero value on the only four lattice points closest to $(p_x,p_y)$, it simplifies Eq.\eqref{eq:bilinear} to a weighted average at the four locations.

\noindent
\textbf{Deformable multi-head attention.} The features $\tilde{x}$ are sampled with bilinear interpolation at the locations of deformed points. Then the deformed keys and values are obtained by the projections $W_k$ and $W_v$, respectively. We summarize the above components along with the queries $q$ as follows:
\begin{align}
    q&=xW_q,\ \tilde{k}=\tilde{x}W_k,\ \tilde{v}=\tilde{x}W_v, \label{eq:proj_dmha}
    \\
    \text{with}\ &\Delta{}p=\theta_\text{offset}(q), \ \tilde{x}=\phi(x;p+\Delta{}p), \label{eq:sampling}
\end{align}
where $\tilde{k}$ and $\tilde{v}$ represent the deformed key and value embeddings. To achieve the deformable multi-head attention, the queries $q$ are attended to the deformed keys $\tilde{k}$, modeling the relations between the query features and the pivot regions of the image. The deformed values $\tilde{v}$ are then aggregated under the guidance of the deformable attention map. The mechanism on the $m$-th attention head is formulated as
\begin{equation}
z^{(m)}=\sigma\left(\cfrac{q^{(m)}\tilde{k}^{(m)\top}}{\sqrt{d}}+\phi(\hat{B};R)\right)\tilde{v}^{(m)},\label{eq:dmha}
\end{equation}
where $\phi(\hat{B};R)\in\mathbb{R}^{HW\times{}H_GW_G}$ corresponds to the relative position bias following previous work while with several adaptations to the deformable locations, which will be discussed later in this section. Finally, the aggregated features of each head $z^{(m)}$ are concatenated together and projected through $W_o$ to get the final output $z$ as Eq.\eqref{eq:MHSA}.

\noindent
\textbf{Deformable relative position bias.} It is worth noting that the integration of position information into attention enhances the performance of models, including APE~\cite{ViT16x16}, RPE~\cite{Swin}, CPE~\cite{cpvt}, LogCPB~\cite{Swinv2}, \textit{etc}. The relative position embedding (RPE) proposed in Swin Transformer encodes the relative positions between every pair of query and key, which augments the vanilla attention with spatial inductive bias. This explicit modeling on relative locations is best suitable for deformable attention, where the deformed keys have arbitrary continuous locations rather than fixed discrete grids. Following RPE, considering a feature map of shape $H\!\times{}\!W$, its relative coordinate displacements lie in the range of $[-H,+H]$ and $[-W,+W]$ at two spatial dimensions. Thus, an RPB table $\hat{B}\in\mathbb{R}^{(2H-1)\times{}(2W-1)}$ holding parameters is constructed to determine the deformable relative position bias by looking up the table. Since our deformable attention has continuous key positions, we compute the relative locations in the normalized range $[-1,+1]$ and then sample $\phi(\hat{B};R)$ with bilinear interpolation in the parameterized bias table $\hat{B}$ by the continuous relative displacements to cover all possible offset values.

\noindent
\textbf{Offset generation.} 
As stated above, a lightweight sub-network is adopted for offset generation, which consumes the query features and outputs the corresponding offset values for each reference point. Considering that each reference point covers a local $r\!\times{}\!r$ region, the offset generation network should also have at least $r\!\times{}\!r$ perception of the local features to learn reasonable offsets. Therefore, we design the sub-network as two convolution modules with a combination of normalization and activation, as depicted in Figure \ref{fig:fig2}(b). The input features are first passed through a $k\!\times \!k$ depth-wise convolution with $r$ stride to capture local features, where $k$ is slightly larger than $r$ to ensure the reasonable receptive field, followed by a LayerNorm and a GELU activation. Then, a $1\!\times{}\!1$ convolution is adopted to predict the 2-D offsets, whose bias is dropped to alleviate the inappropriate compulsive shift for all locations. 

\noindent
\textbf{Offset groups.} 
To promote the diversity of the deformed points, we follow a similar paradigm in MHSA that the token channels are split into multiple heads to compute a variety of attention. Therefore, we split the channels into $G$ groups to generate diverse offsets. The offset generation network shares weights for features from different groups to generate the corresponding offsets. In practice, the head number $M$ of attention is set to be multiple times the size of the offset groups $G$, guaranteeing that multiple attention heads are assigned to one group of sampled features.

\subsubsection{Complexity Analysis}\label{sec:m_complex}

Deformable multi-head attention (DMHA) has a similar computation cost as the counterpart in PVT or Swin Transformer. The only additional overheads come from the offset generation network and the bilinear feature sampling. The time complexity of DMHA can be summarized as:
\begin{equation}
\Omega(\text{DMHA})\!=\!\underbrace{2HWN_\text{s}C\!\!+\!\!2HWC^2\!\!+\!\!2N_\text{s}C^2}_{\text{multi-head self-attention}}\!+\!\underbrace{(k^2\!+\!6)N_\text{s}C}_{\text{\textcolor{blue}{offsets~\&~sampling}}},\label{eq:time_complexity}
\end{equation}
where $N_\text{s}\!=\!H_GW_G\!=\!HW/r^2$ is the number of sampled points. It can be immediately seen that the computational cost of the offset network has linear complexity \textit{w.r.t.} the spatial and channel size, which is comparably minor to the cost for attention computation. Besides, the smaller $N_\text{s}$ further reduces the computation cost in the projection layers of keys and values by first sampling the features. 

We adopt Floating Point Operations (FLOPs) to measure the computational complexity. \textit{E.g.}, the $3^\text{rd}$ stage of a hierarchical model with $224\!\times{}\!224$ input for image classification usually has the computation sizes of $H\!=\!W\!=\!14$, $N_\text{s}\!=\!49$, $C\!=\!384$, thus the computational complexity for multi-head self-attention in a single DMHA block is 79.63M in FLOPs. Combining the offset generation network, whose additional overhead is 0.58M in FLOPs, which is only $0.7\%$ in 80.21 MFLOPs of the whole module. Additionally, the complexity could be further reduced by enlarging downsample factor $r$, making it friendly to tasks with much higher resolution inputs such as object detection and instance segmentation.

\subsection{Model Architectures}\label{sec:m_arch}

\begin{figure}[t]
    \begin{center}
    \includegraphics[width=\linewidth]{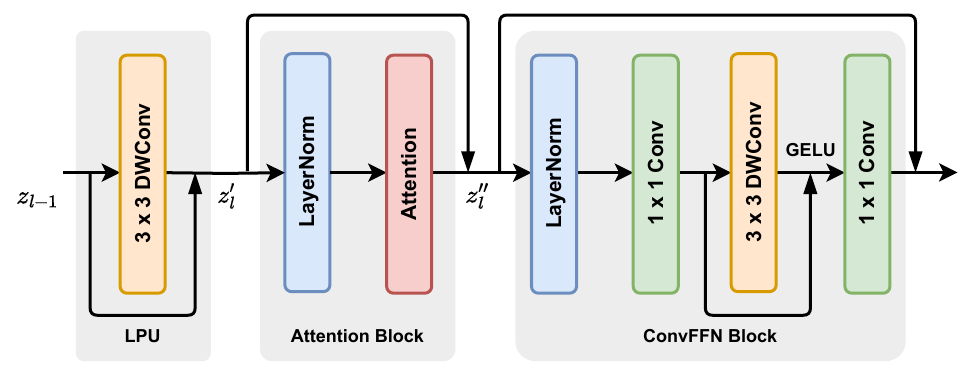}
    \vspace{-0.5cm}
    \caption{Detailed architecture of each Transformer block in \ourmethod{}++, enhanced with Local Perception Unit (LPU) and ConvFFN blocks.}
    \vspace{-0.5cm}
    \label{fig:fig_block}
    \end{center}
\end{figure}

\subsubsection{Building \omlong{}}\label{sec:m_build_dat}

We replace the vanilla MHSA (Eq.\eqref{eq:attn}) with the proposed DMHA (Eq.\eqref{eq:dmha}) in the attention layer (Eq.\eqref{eq:att_layer}) and combine it with an MLP (Eq.\eqref{eq:mlp_layer}) to build a Vision Transformer block with deformable attention. Based on these blocks, our model, \textbf{\omlong{} (\ourmethod{})}, shares a similar pyramid structure with~\cite{PVT,Swin,dpt,botnet}, which is broadly applicable to various vision tasks requiring multiscale feature maps. The architecture of \ourmethod{} is illustrated in Figure~\ref{fig:fig_model_arch}, in which an input image with shape $H\!\times{}\!W\!\times{}\!3$ is firstly embedded by a convolutional stem with stride 4 to encode the $H/4\!\times{}\!W/4\times{}\!C$ patch embeddings. Aiming to build a hierarchical feature pyramid, the backbone includes 4 model stages with a progressively increasing stride at each stage. Between two consecutive stages, there is a convolutional downsampling layer with stride 2. In each stage, sequential building blocks of different types of attention extract features from the patch embeddings by spatial mixing in the attention layers and channel mixing in the MLP layers.

After stages of processing, \ourmethod{} learns the representations of the input image hierarchically, thus ready for various vision tasks. For the classification task, we first normalize the output from the last stage by LN~\cite{LN} and then adopt a linear classifier with pooled features to predict the class label. For the object detection, instance segmentation, and semantic segmentation tasks, \ourmethod{} plays the role of a feature extractor in some well-designed detectors or segmentors. To build the feature pyramid for these tasks, we add an LN layer to the features from each stage before feeding them into the following modules such as FPN~\cite{fpn} for object detection or decoders in semantic segmentation. 

\begin{figure}[t]
    \centering
    \includegraphics[width=\linewidth]{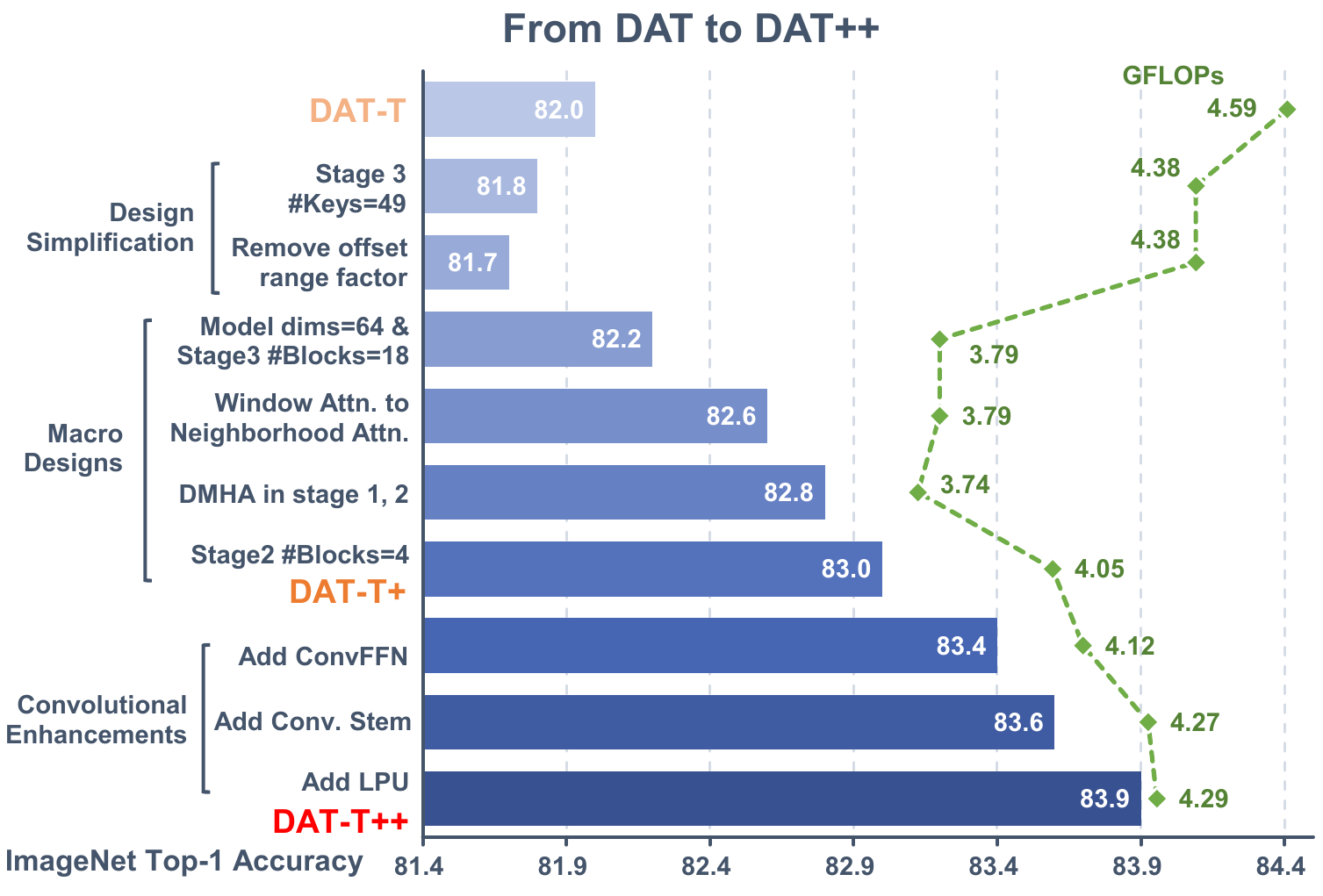}
    \vspace{-0.2in}
    \caption{Roadmap from \ourmethod{}~\cite{Xia_2022_CVPR} to the improved \ourmethod{}++. From the top to the bottom, each row represents a version of model, where the modifications are appended to the model in the last row. The blue bar chart reports the top-1 accuracy of each model on ImageNet next to the green line chart which shows the complexity of each model in GFLOPs.}
    \label{fig:roadmap}
    \vspace{-0.3cm}
\end{figure}

\subsubsection{Enhancing \ourmethod{} to \ourmethod{}++}
\label{sec:m_conv}

To fully explore the capacity of \ourmethod{}, we consider incorporating several enhancements of learning better local features into the architecture to boost performance and build an improved version named \ourmethod{}++, which is sketched as a roadmap in Figure~\ref{fig:roadmap}. We introduce these enhancements as follows, and detailed discussions are provided in Sec.~\ref{sec:e_abl}.

\noindent
\textbf{Overlapped patch embedding layers} provide remarkable gains over their non-overlapped counterparts~\cite{ViT16x16,deit,cait,Swin,PVT}, which are adopted in \ourmethod{}++ to take the place of the stem layers and downsampling layers, as done in~\cite{biformer,svit,cmt,DCNv3}. This overlapped convolution manner could mitigate the severe information loss in the unfolding operation and produce high-quality feature maps, which is essential to learn accurate offsets for DMHA to dynamically focus on the important regions in the image. Specifically, we use the $3\!\times{}\!3$ convolution with stride 2 to embed image pixels into tokens, and downsample feature maps across stages.

\begin{table}
\newcommand{\tabincell}[2]{\begin{tabular}{@{}#1@{}}#2\end{tabular}}
\begin{center}
\setlength{\tabcolsep}{0.9mm}{
\renewcommand\arraystretch{1.2}
\caption{\ourmethod{}++ model architecture specifications. $N_i$: Number of blocks at stage $i$. $C$: Base channels in each block. $r$: Downsample ratio of deformed points. $M$: Number of attention heads in DMHA. $G$: Number of offset groups in DMHA. $K$: Kernel size of local neighborhood attention. The spatial size and stride of feature maps are labeled under each stage.}
\label{tab:arch}
\vspace{-0.2cm}
\resizebox{\linewidth}{!}{
\begin{tabular}{c|c|c|c}
\thickhline
\multicolumn{4}{c}{\textbf{Variant Architectures of \omlong{}}} \\
 ($224\!\times{}\!224$) & \ourmethod{}-T++ & \ourmethod{}-S++ & \ourmethod{}-B++ \\
\hline
Stage 1 & $N_1\!=\!1,C\!=\!64$ & $N_1\!=\!1,C\!=\!96$ & $N_1\!=\!1,C\!=\!128$ \\
($56\!\times{}\!56$) & $r\!=\!8,M\!=\!2$ & $r\!=\!8,M\!=\!3$ & $r\!=\!8,M\!=\!4$ \\
Stride: 4 & $G\!=\!1,K\!=\!7$ & $G\!=\!1,K\!=\!7$ & $G\!=\!2,K\!=\!7$ \\
\hline
Stage 2 & $N_2\!=\!2,C\!=\!128$ & $N_2\!=\!2,C\!=\!192$ & $N_2\!=\!2,C\!=\!256$ \\
($28\!\times{}\!28$) & $r\!=\!4,M\!=\!4$ & $r\!=\!4,M\!=\!6$ & $r\!=\!4,M\!=\!8$ \\
Stride: 8 & $G\!=\!2,K\!=\!7$ & $G\!=\!2,K\!=\!7$ & $G\!=\!4,K\!=\!7$ \\
\hline
Stage 3 & $N_3\!=\!9,C\!=\!256$ & $N_3\!=\!9,C\!=\!384$ & $N_3\!=\!9,C\!=\!512$ \\
($14\!\times{}\!14$) & $r\!=\!2,M\!=\!8$ & $r\!=\!2,M\!=\!12$ & $r\!=\!8,M\!=\!16$\\
Stride: 16 & $G\!=\!4,K\!=\!7$ &$G\!=\!3,K\!=\!7$ & $G\!=\!8,K\!=\!7$ \\
\hline
Stage 4 & $N_4\!=\!1,C\!=\!512$ & $N_4\!=\!1,C\!=\!768$ & $N_4\!=\!1,C\!=\!1024$ \\
($7\!\times{}\!7$) & $r\!=\!1,M\!=\!16$ & $r\!=\!1,M\!=\!24$ & $r\!=\!1,M\!=\!32$ \\
Stride: 32 & $G\!=\!8$ & $G\!=\!6$ & $G\!=\!16$ \\
\thickhline
\end{tabular}}
}
\vspace{-0.2in}
\end{center}
\end{table}

\noindent
\textbf{Local Perception Unit (LPU)}~\cite{cmt} is a depth-wise convolution wrapped by a residual connection. Similar to CPE~\cite{cpvt}, this layer is usually placed at the top of every transformer block to enhance positional information implicitly. Since deformable attention is highly sensitive to positional features and the geometric distributions in the image, the implicit conditional position encoding added to feature maps could provide hints of the important regions, which benefits the learning of the offset generation module of DMHA.

\noindent
\textbf{ConvFFN} adds a depth-wise convolution between two fully connected layers in the FFN (MLP) for channel mixing, reinforcing the spatial modeling ability ~\cite{PVTv2,cmt,hilo,replknet}. ConvFFN serves as a local feature modeling operation, which complements the global attention DMHA with better locality modeling. \ourmethod{}++ chooses a variant of ConvFFN with a skip connection rather than plain convolution in PVTv2~\cite{PVTv2} or additional BatchNorms~\cite{BN} in CMT~\cite{cmt}.

The building block of \ourmethod{}++ equipped with the above enhancements is illustrated in Figure~\ref{fig:fig_model_arch}~-~\ref{fig:fig_block}. In \ourmethod{}++, we modify the $l$-th Transformer encoder block (Eq.\eqref{eq:att_layer}~-~\eqref{eq:mlp_layer}) with convolutions injected:
\begin{align}
    z_l'&=\text{LPU}\left(z_{l-1}\right), \label{eq:lpu_layer} \\
    z_l''&=\text{MHSA}\left(\text{LN}(z_l')\right)+z_l', \label{eq:att_layer2} \\
    z_l&=\text{ConvFFN}\left(\text{LN}(z_l'')\right)+z_l''.\label{eq:mlp_layer2}
\end{align}
To make fair comparisons, we carefully ablate these additional components in Sec.~\ref{sec:e_abl} to verify the effectiveness of our proposed deformable attention.

\subsubsection{Architecture Design}\label{sec:m_alter_block}

We introduce successive local attention and deformable attention blocks in the first three stages of \ourmethod{}++. Unlike the conference version~\cite{Xia_2022_CVPR}, which leverages window-based attention in Swin Transformer~\cite{Swin} as the local attention layer, we opt for Neighborhood Attention~\cite{nat} as its efficient local self-attention possesses more representation power.

In the first three stages of \ourmethod{}++, the feature maps are processed by a transformer block with local attention to aggregate information within a neighbor window for visual tokens and then pass through the deformable attention block to model the global relations between the locally augmented tokens. This alternate design of attention blocks with local and global receptive fields helps the model learn strong representations, sharing similar thoughts in~\cite{twins,pointformer,glit,TNT,scalablevit,dinat,gcvit}, which has been shown to be an effective design. Since the spatial resolution is rather limited in the last stage, the neighbor size of the local attention is close to the size of feature maps. The local attention performs similarly to the vanilla MHSA. Therefore, all the transformer blocks in the fourth stage of \ourmethod{}++ are set to deformable attention. We discuss this design in Sec.~\ref{sec:e_abl}.

We build three variants of \ourmethod{}++ at different scales of parameters and FLOPs for fair comparisons with other ViTs and CNNs. We change the model scale by increasing the hidden dimensions, leading to \ourmethod{}-T/S/B++. The detailed architectures are summarized in Table~\ref{tab:arch}. Note that there are other design choices for the local modeling layers, \textit{e.g.}, large kernel depth-wise convolutions~\cite{convnext}, and window-based local attention~\cite{Swin}. We show that DMHA is versatile and works well with other local operators in Sec.~\ref{sec:e_abl}.

\section{Experiments}\label{sec:exps}

We analyze~\ourmethod{}++ on several representative visual recognition tasks, including ImageNet~\cite{in1k} image classification, MS-COCO~\cite{coco} object detection, and ADE20K~\cite{ade20k} semantic segmentation. In addition to comparing~\ourmethod{} with other leading models on these standard benchmark datasets, we provide detailed ablation experiments and visualization to analyze the effectiveness of the proposed~\ourmethod{}++.

\subsection{Image Classification}\label{sec:e_m_cls}

\begin{table}[t]
\begin{center}
\caption{Detailed hyper-parameters for training variants of \ourmethod{}++ on ImageNet. PT and FT means pretraining and finetuning, respectively.}
\label{tab:impl}
\vspace{-0.2in}
\setlength{\tabcolsep}{0.5mm}{
\renewcommand\arraystretch{1.2}
\resizebox{\linewidth}{!}{
\begin{tabular}{l|cccc}
\thickhline
Settings & \ourmethod{}-T++ & \ourmethod{}-S++ & \multicolumn{2}{c}{\ourmethod{}-B++} \\
\cline{2-4}
\hline
Input resolution & \textsf{PT} 224$^2$ & \textsf{PT} 224$^2$ & \textsf{PT} 224$^2$ &  \textsf{FT} 384$^2$ \\
Batch size & 4096 & 4096 & 4096 & 512 \\
Optimizer & AdamW & AdamW & AdamW & AdamW \\
Learning rate & 4$\times10^{-3}$ & 4$\times10^{-3}$ & 4$\times10^{-3}$ & 2$\times10^{-5}$ \\
LR schedule & cosine & cosine & cosine & cosine \\
Weight decay & 5$\times10^{-2}$ & 5$\times10^{-2}$ & 5$\times10^{-2}$ & 1$\times10^{-8}$ \\
Warmup epochs & 5 & 5 & 5 & 0 \\
Epochs & 300 & 300 & 300 & 30 \\
\hline
Horizontal flip & \ding{51} & \ding{51} & \ding{51} & \ding{51} \\
Random resize \& Crop & \ding{51} & \ding{51} & \ding{51} & \ding{51} \\
AutoAugment & \ding{51} & \ding{51} & \ding{51} & \ding{51} \\
Mixup alpha & 0.8 & 0.8 & 0.8 & 0.8 \\ 
Cutmix alpha & 1.0 & 1.0 & 1.0 & 1.0 \\
Random erasing prob. & 0.25 & 0.25 & 0.25 & 0.25 \\
Color jitter  & 0.4 & 0.4 & 0.4 & 0.4 \\
\hline
Label smoothing & 0.1 & 0.1 & 0.1 & 0.1 \\
Dropout & \ding{55} & \ding{55} &  \ding{55} & \ding{55} \\
Droppath rate & 0.2 & 0.4 & 0.6 & 0.6 \\
Repeated augment & \ding{55} & \ding{55} & \ding{55} & \ding{55} \\
Grad. clipping & 5.0  & 5.0 & 5.0 & 5.0 \\
\thickhline
\end{tabular}}}
\vspace{-0.5cm}
\end{center}
\end{table}

\begin{table}[htbp]
\newcommand{\tabincell}[2]{\begin{tabular}{@{}#1@{}}#2\end{tabular}}
\begin{center}
\caption{Image classification performance of \ourmethod{} and other popular vision backbones on ImageNet-1K validation set. The models are divided into Tiny, Small, and Base groups by computational complexity.
}\label{tab:in1k}
\vspace{-0.1in}
\setlength{\tabcolsep}{0.9mm}{
\renewcommand\arraystretch{1.25}
\resizebox{\linewidth}{!}{
\begin{tabular}{c|l|ccc|c}
\thickhline
\multicolumn{6}{c}{\textbf{ImageNet-1K Classification with ImageNet-1K training}} \\
Scale & Method & Resolution & FLOPs & \#Param & Top-1 Acc.\\
\hline
\multirow{17}{*}{Tiny} & DeiT-S~\cite{deit} & $\text{224}^\text{2}$ & 4.6G & 22M & 79.8 \\
&PVT-S~\cite{PVT} & $\text{224}^\text{2}$ & 3.8G & 25M & 79.8 \\
&DPT-S~\cite{dpt} & $\text{224}^\text{2}$ & 4.0G & 26M & 81.0 \\
&Swin-T~\cite{Swin} &  $\text{224}^\text{2}$ & 4.5G & 29M & 81.3 \\
&PVTv2-B2~\cite{PVTv2} & $\text{224}^\text{2}$ & 4.0G & 25M & 82.0 \\
&ConvNeXt-T~\cite{convnext} & $\text{224}^\text{2}$ & 4.5G & 29M & 82.1 \\
&Focal-T~\cite{focal} & $\text{224}^\text{2}$ & 4.9G & 29M & 82.2 \\
&ViL-S~\cite{vil} & $\text{224}^\text{2}$ & 4.9G & 25M & 82.4 \\
&DiNAT-T~\cite{dinat} & $\text{224}^\text{2}$ &  4.3G & 28M & 82.7 \\
&CSWin-T~\cite{cswin} & $\text{224}^\text{2}$ & 4.3G & 23M & 82.7 \\
&HorNet-T~\cite{hornet} & $\text{224}^\text{2}$ & 3.9G & 23M & 83.0 \\
&NAT-T~\cite{nat} & $\text{224}^\text{2}$ & 4.3G & 28M & 83.2 \\
&RegionViT-S+~\cite{rgvit} & $\text{224}^\text{2}$ & 5.7G & 31M & 83.3 \\
&InternImage-T~\cite{DCNv3} & $\text{224}^\text{2}$ & 5.0G & 30M & 83.5 \\
&BiFormer-S~\cite{biformer} & $\text{224}^\text{2}$ & 4.5G & 26M & 83.8 \\
& \cellcolor{mygray}\ourmethod{}-T~\cite{Xia_2022_CVPR} & \cellcolor{mygray}$\text{224}^\text{2}$ & \cellcolor{mygray}4.6G & \cellcolor{mygray}29M & \cellcolor{mygray}82.0 \\
& \cellcolor{mygray}\textbf{\ourmethod{}-T++} & \cellcolor{mygray}$\text{224}^\text{2}$ & \cellcolor{mygray}4.3G & \cellcolor{mygray}24M & \cellcolor{mygray}\textbf{83.9} \\
\hline
\multirow{17}{*}{Small} & PVT-M \cite{PVT} & $\text{224}^\text{2}$ & 6.9G & 46M & 81.2 \\
&PVT-L \cite{PVT} & $\text{224}^\text{2}$ & 9.8G & 61M & 81.7 \\
&DPT-M \cite{dpt} & $\text{224}^\text{2}$ & 6.9G & 46M & 81.9 \\
&Swin-S \cite{Swin} &  $\text{224}^\text{2}$ & 8.8G & 50M & 83.0 \\
&ConvNeXt-S~\cite{convnext} & $\text{224}^\text{2}$ & 8.7G & 50M & 83.1 \\
&RegionViT-M+~\cite{rgvit} & $\text{224}^\text{2}$ & 7.9G & 42M & 83.4 \\
&ViL-M~\cite{vil} & $\text{224}^\text{2}$ & 8.7G & 40M & 83.5 \\
&PVTv2-B4~\cite{PVTv2} & $\text{224}^\text{2}$ & 10.1G & 62M & 83.6 \\
&Focal-S~\cite{focal} & $\text{224}^\text{2}$ & 9.4G & 51M & 83.6 \\
&CSWin-S~\cite{cswin} & $\text{224}^\text{2}$ & 6.9G & 35M & 83.6 \\
&NAT-S~\cite{nat} & $\text{224}^\text{2}$ & 7.8G & 51M & 83.7 \\
&DiNAT-S~\cite{dinat} & $\text{224}^\text{2}$ & 7.8G & 51M & 83.8 \\
&HorNet-S~\cite{hornet} & $\text{224}^\text{2}$ & 8.7G & 50M & 84.0 \\
&InternImage-S~\cite{DCNv3} & $\text{224}^\text{2}$ & 8.0G & 50M & 84.2 \\
&BiFormer-B~\cite{biformer} & $\text{224}^\text{2}$ & 9.8G & 57M & 84.3 \\
&\cellcolor{mygray}\ourmethod{}-S~\cite{Xia_2022_CVPR} & \cellcolor{mygray}$\text{224}^\text{2}$ & \cellcolor{mygray}9.0G & \cellcolor{mygray}50M & \cellcolor{mygray}83.7 \\
&\cellcolor{mygray}\textbf{\ourmethod{}-S++} & \cellcolor{mygray}$\text{224}^\text{2}$ & \cellcolor{mygray}9.4G & \cellcolor{mygray}53M & \cellcolor{mygray}\textbf{84.6}\\
\hline
\multirow{21}{*}{Base} & DeiT-B~\cite{deit} & $\text{224}^\text{2}$ & 17.5G & 86M & 81.8 \\
&Swin-B \cite{Swin} & $\text{224}^\text{2}$ & 15.5G & 88M & 83.5 \\
&ViL-B~\cite{vil} & $\text{224}^\text{2}$ & 13.4G & 56M & 83.7 \\
&RegionViT-B+~\cite{rgvit} & $\text{224}^\text{2}$ & 13.6G & 74M & 83.8 \\
&ConvNeXt-B~\cite{convnext} & $\text{224}^\text{2}$ & 15.4G & 89M & 83.8 \\
&PVTv2-B5~\cite{PVTv2} & $\text{224}^\text{2}$ & 11.8G & 82M & 83.8 \\
&Focal-B~\cite{focal} & $\text{224}^\text{2}$ & 16.4G & 90M & 84.0 \\
&CSWin-B~\cite{cswin} & $\text{224}^\text{2}$ & 15.0G & 78M & 84.2 \\
&HorNet-B~\cite{hornet} & $\text{224}^\text{2}$ & 15.5G & 88M & 84.3 \\
&NAT-B~\cite{nat} & $\text{224}^\text{2}$ & 13.7G & 90M & 84.3 \\
&DiNAT-B~\cite{dinat} & $\text{224}^\text{2}$ & 13.7G & 90M & 84.4 \\
&InternImage-B~\cite{DCNv3} & $\text{224}^\text{2}$ & 16.0G & 97M & \textbf{84.9} \\
&\cellcolor{mygray}\ourmethod{}-B~\cite{Xia_2022_CVPR} & \cellcolor{mygray}$\text{224}^\text{2}$ & \cellcolor{mygray}15.8G & \cellcolor{mygray}88M & \cellcolor{mygray}84.0\\
&\cellcolor{mygray}\textbf{\ourmethod{}-B++} & \cellcolor{mygray}$\text{224}^\text{2}$ & \cellcolor{mygray}16.6G & \cellcolor{mygray}93M & \cellcolor{mygray}\textbf{84.9}\\
\cline{2-6}
&DeiT-B~\cite{deit} & $\text{384}^\text{2}$ & 55.4G & 86M & 83.1 \\
&Swin-B~\cite{Swin} & $\text{384}^\text{2}$ & 47.2G & 88M & 84.5 \\
&ConvNeXt-B~\cite{convnext} & $\text{384}^\text{2}$ & 45.0G & 89M & 85.1 \\
&CSWin-B~\cite{cswin} & $\text{384}^\text{2}$ & 47.0G & 78M & 85.4 \\
&HorNet-B~\cite{hornet} & $\text{384}^\text{2}$ & 45.2G & 92M & 85.6 \\
&\cellcolor{mygray}\ourmethod-B~\cite{Xia_2022_CVPR} & \cellcolor{mygray}$\text{384}^\text{2}$ & \cellcolor{mygray}49.8G & \cellcolor{mygray}88M & \cellcolor{mygray}84.8 \\
&\cellcolor{mygray}\textbf{\ourmethod-B++} & \cellcolor{mygray}$\text{384}^\text{2}$ & \cellcolor{mygray}49.7G & \cellcolor{mygray}94M & \cellcolor{mygray}\textbf{85.9} \\
\thickhline
\end{tabular}}}
\end{center}
\end{table}

\begin{table}[t]
\begin{center}
\caption{Results on MS-COCO object detection with RetinaNet~\cite{rtn}. This table shows the number of parameters, computational cost (FLOPs), mAP at different mIoU thresholds and object sizes. The FLOPs are computed over the entire detector at the resolution of 1280$\times$800$\times$3.}
\label{tab:det_retinanet}
\vspace{-0.1in}
\setlength{\tabcolsep}{0.3mm}{
\renewcommand\arraystretch{1.3}
\resizebox{\linewidth}{!}{
\begin{tabular}{l|c|c|c|ccc|ccc}
\thickhline
\multicolumn{10}{c}{\textbf{RetinaNet Object Detection on MS-COCO}} \\
Method & FLOPs & \#Param & Sch. & AP & AP$_\text{50}$ & AP$_\text{75}$ & AP$_{s}$ & AP$_{m}$ & AP$_{l}$ \\
\hline
PVT-S~\cite{PVT}& 286G & 34M & 1x & 40.4 & 61.3 & 43.0 & 25.0 & 42.9 & 55.7 \\
Swin-T~\cite{Swin}& 248G & 38M & 1x & 41.7 & 63.1 & 44.3 & 27.0 & 45.3 & 54.7 \\
Focal-T~\cite{focal}& 265G & 39M & 1x & 43.7 & - & - & - & - & - \\
RegionViT-S+~\cite{rgvit}& 205G & 42M & 1x & 43.9 & 65.5 & 47.3 & 28.5 & 47.3 & 58.0 \\
ViL-S~\cite{vil}& 255G & 36M & 1x & 44.2 & 65.2 & 47.6 & 28.8 & 48.0 & 57.8 \\
PVTv2-B2~\cite{PVTv2}& 291G & 35M & 1x & 44.6 & 65.6 & 47.6 & 27.4 & 48.8 & 58.6 \\
\rowcolor{mygray}
\ourmethod-T~\cite{Xia_2022_CVPR} & 253G & 38M & 1x & 42.8 & 64.4 & 45.2 & 28.0 & 45.8 & 57.8 \\
\rowcolor{mygray}
\textbf{\ourmethod{}-T++} & 283G & 34M & 1x & \textbf{46.8} & 68.4 & 50.3 & 30.8 & 51.9 & 62.5 \\
\hline
PVT-S~\cite{PVT} & 286G & 34M & 3x & 42.3 & 63.1 & 44.8 & 26.7 & 45.1 & 57.2 \\
Swin-T~\cite{Swin} & 248G & 38M & 3x & 44.8 & 66.1 & 48.0 & 29.2 & 48.6 & 58.6 \\
Focal-T~\cite{focal}& 265G & 39M & 3x & 45.5 & 66.3 & 48.8 & 31.2 & 49.2 & 58.7 \\
ViL-S~\cite{vil}& 255G & 36M & 3x & 45.9 & 66.6 & 49.0 & 30.9 & 49.3 & 59.9 \\
RegionViT-S+~\cite{rgvit}& 205G & 42M & 3x & 46.7 & 68.2 & 50.2 & 30.8 & 50.8 & 62.4 \\
\rowcolor{mygray}
\ourmethod-T{}~\cite{Xia_2022_CVPR} & 253G & 38M & 3x & 45.6 & 67.2 & 48.5 & 31.3 & 49.1 & 60.8 \\
\rowcolor{mygray}
\textbf{\ourmethod-T++} & 283G & 34M & 3x & \textbf{49.2} & 70.3 & 53.0 & 32.7 & 53.4 & 64.7 \\
\hline
PVT-L~\cite{PVT} & 475G & 71M & 1x & 42.6 & 63.7 & 45.4 & 25.8 & 46.0 & 58.4 \\
Swin-S~\cite{Swin} & 339G & 60M & 1x & 44.5 & 66.1 & 47.4 & 29.8 & 48.5 & 59.1 \\
RegionViT-B+~\cite{rgvit} & 328G & 85M & 1x & 44.6 & 66.4 & 47.6 & 29.6 & 47.6 & 59.0 \\
Focal-S~\cite{focal}& 367G & 62M & 1x & 45.6 & - & - & - & - & - \\
PVTv2-B4~\cite{PVTv2} & 482G & 72M & 1x & 46.1 & 66.9 & 49.2 & 28.4 & 50.0 & 62.2 \\
PVTv2-B5~\cite{PVTv2} & 539G & 92M & 1x & 46.2 & 67.1 & 49.5 & 28.5 & 50.0 & 62.5 \\
Focal-B~\cite{focal} & 514G & 101M & 1x & 46.3  & - & - & - & - & - \\
ViL-M~\cite{vil}& 330G & 51M & 1x & 46.8 & 68.1 & 50.0 & 31.4 & 50.8 & 60.8 \\
ViL-B~\cite{vil}& 421G & 67M & 1x & 47.8 & 69.2 & 51.4 & 32.4 & 52.3 & 61.8 \\
\rowcolor{mygray}
\ourmethod-S~\cite{Xia_2022_CVPR} & 359G & 60M & 1x & 45.7 & 67.7 & 48.5 & 30.5 & 49.3 & 61.3 \\
\rowcolor{mygray}
\textbf{\ourmethod-S++} & 410G & 63M & 1x & \textbf{48.3} & 70.0 & 51.8 & 32.3 & 52.4 & 63.1 \\
\hline
Focal-B~\cite{focal} & 514G & 101M & 3x & 46.9 & 67.8 & 50.3 & 31.9 & 50.3 & 61.5 \\
RegionViT-B+~\cite{rgvit} & 328G & 85M & 3x & 46.9 & 68.3 & 50.3 & 31.1 & 50.5 & 62.4 \\
Focal-S~\cite{focal}& 367G & 62M & 3x & 47.3 & 67.8 & 51.0 & 31.6 & 50.9 & 61.1 \\
Swin-S~\cite{Swin} & 339G & 60M  & 3x & 47.3 & 68.6 & 50.8 & 31.9 & 51.8 & 62.1 \\
ViL-M~\cite{vil}& 330G & 51M & 3x & 47.9 & 68.8 & 51.3 & 32.4 & 51.9 & 61.8 \\
ViL-B~\cite{vil}& 421G & 67M & 3x & 48.6 & 69.4 & 52.2 & 34.1 & 52.5 & 61.9 \\
\rowcolor{mygray}
\ourmethod-S~\cite{Xia_2022_CVPR} & 359G & 60M & 3x & 47.9 & 69.6 & 51.2 & 32.3 & 51.8 & 63.4 \\
\rowcolor{mygray}
\textbf{\ourmethod-S++} & 410G & 63M & 3x & \textbf{50.2} & 71.5 & 54.0 & 34.7 & 54.6 & 65.3 \\
\thickhline
\end{tabular}}}
\vspace{-0.6cm}
\end{center}
\end{table}

\noindent
\textbf{Settings.}
ImageNet~\cite{in1k} is a widely used image classification dataset for deep neural networks, which contains 1.28M examples for training and 50K examples for validation. We train three variants of~\ourmethod{}++, including~\ourmethod{}-T/S/B++, on the training split and report the Top-1 accuracy on the validation split to compare with other Vision Transformers and Convolution Neural Networks. 

Following common practices in ~\cite{deit,Swin,cswin,convnext,hornet,DCNv3}, we train~\ourmethod{}-T/S/B++ on ImageNet-1K for 300 epochs. For all model variants, we use AdamW \cite{adamw} optimizer with a cosine learning rate decay in the pretraining phase. The data augmentation configurations follow DeiT~\cite{deit} and Swin Transformer~\cite{Swin}, with RandAugment~\cite{randaug}, Mixup~\cite{mixup}, CutMix~\cite{cutmix}, Random Erasing~\cite{randerase}, \textit{etc.} Without using vanilla dropout~\cite{dropout}, we choose stochastic depth~\cite{sdepth} to regularize the training and avoid overfitting. Table~\ref{tab:impl} lists the detailed hyper-parameters, such as learning rates warm-up and gradient clipping, which are adopted to stabilize training.

We follow the common practices in many previous works~\cite{Swin,convnext} and mainly tune the drop path rate to avoid overfitting. We do not involve any other techniques, such as EMA~\cite{ema}, LayerScale~\cite{cait}, and layer-wise learning rate decay~\cite{convnext,hornet}, which do not induce any obvious improvements in our early experiments.

\begin{table*}[t]
\begin{center}
\caption{Object detection and instance segmentation performance on MS-COCO dataset with Mask R-CNN~\cite{mrcn} detector. This table shows the number of parameters, computational cost (FLOPs), mAP at different IoU thresholds, and different object sizes. The FLOPs are computed over the entire detector at the resolution of 1280$\times$800$\times$3. The results of ConvNeXt~\cite{convnext} at Tiny-1x, Small-1x, and Small-3x are copied from~\cite{DCNv3}.}\label{tab:det_mrcn}
\vspace{-0.2in}
\setlength{\tabcolsep}{1.2mm}{
\renewcommand\arraystretch{1.2}
\resizebox{\linewidth}{!}{
\begin{tabular}{l|c|c|c|ccc|ccc|ccc|ccc}
\thickhline
\multicolumn{16}{c}{\textbf{Mask R-CNN Object Detection \& Instance Segmentation on MS-COCO}} \\
Method & FLOPs & \#Param & Schedule & AP$^b$ & AP$^b_\text{50}$ & AP$^b_\text{75}$ & AP$^b_s$ & AP$^b_m$ & AP$^b_l$ & AP$^m$ & AP$^m_\text{50}$ & AP$^m_\text{75}$ & AP$^m_s$ & AP$^m_m$ & AP$^m_l$ \\
\hline
Swin-T~\cite{Swin} & 267G & 48M & 1x & 43.7 & 66.6 & 47.7 & 28.5 & 47.0 & 57.3 & 39.8 & 63.3 & 42.7 & 24.2 & 43.1 & 54.6 \\
ConvNeXt-T~\cite{convnext}& 262G & 48M & 1x & 44.2 & 66.6 & 48.3 & - & - & - & 40.1 & 63.3 & 42.8 & - & - & - \\
RegionViT-S+~\cite{rgvit} & 183G & 51M & 1x & 44.2 & 67.3 & 48.2 & 27.5 & 47.4 & 57.8 & 40.9 & 64.1 & 44.0 & 21.6 & 43.4 & 58.9  \\
PVTv2-B2~\cite{PVTv2} & 309G & 45M & 1x & 45.3 & 67.1 & 49.6 & 28.8 & 48.4 & 59.5 & 41.2 & 64.2 & 44.4 & 22.0 & 43.7 & 59.4 \\
CSWin-T~\cite{cswin} & 279G & 42M & 1x & 46.7 & 68.6 & 51.3 & - & - & - & 42.2 & 65.6 & 45.4 & - & - & - \\
InternImage-T~\cite{DCNv3} & 270G & 49M & 1x & 47.2 & 69.0 & 52.1 & 30.9 & 50.8 & 61.7 & 42.5 & 66.1 & 45.8 & 22.7 & 45.7 & 61.2 \\
\rowcolor{mygray}
DAT-T~\cite{Xia_2022_CVPR} & 272G & 48M & 1x & 44.4 & 67.6 & 48.5 & 28.3 & 47.5 & 58.5 & 40.4 & 64.2 & 43.1 & 23.9 & 43.8 & 55.5 \\ 
\rowcolor{mygray}
\textbf{\ourmethod-T++} & 301G & 45M & 1x & \textbf{48.7} & 70.9 & 53.7 & 32.8 & 52.4 & 63.5 & \textbf{43.7} & 67.9 & 47.3 & 24.5 & 47.4 & 62.4 \\
\hline
Swin-T~\cite{Swin} & 267G & 48M & 3x & 46.0 & 68.1 & 50.3 & 31.2 & 49.2 & 60.1 & 41.6 & 65.1 & 44.9 & 25.9 & 45.1 & 56.9 \\
ConvNeXt-T~\cite{convnext} & 262G & 48M & 3x & 46.2 & 67.9 & 50.8 & 29.8 & 49.5 & 59.3 & 41.7 & 65.0 & 45.0 & 21.9 & 44.8 & 59.8 \\
RegionViT-S+~\cite{rgvit} & 183G & 51M & 3x & 47.6 & 70.0 & 52.0 & 31.5 & 51.4 & 62.6 & 43.4 & 67.1 & 47.0 & 24.3 & 46.4 & 62.0  \\ 
NAT-T~\cite{nat} & 258G & 48M & 3x & 47.8 & 69.0 & 52.6 & 32.2 & 51.3 & 61.8 & 42.6 & 66.0 & 45.9 & 23.1 & 45.7 & 61.4 \\
DiNAT-T~\cite{dinat} & 258G & 48M & 3x & 48.6 & 70.3 & 53.5 & 32.5 & 52.4 & 63.3 & 43.5 & 67.2 & 46.7 & 24.2 & 46.9 & 62.3 \\
CSwin-T~\cite{cswin} & 279G & 42M & 3x & 49.0 & 70.7 & 53.7 & - & - & - & 43.6 & 67.9 & 46.6 & - & - & - \\
InternImage-T~\cite{DCNv3} & 270G & 49M & 3x & 49.1 & 70.4 & 54.1 & 33.1 & 52.8 & 63.8 & 43.7 & 67.3 & 47.3 & 24.3 & 47.0 & 62.5 \\
\rowcolor{mygray}
\ourmethod-T~\cite{Xia_2022_CVPR} & 272G & 48M & 3x & 47.1 & 69.2 & 51.6 & 32.0 & 50.3 & 61.0 & 42.4 & 66.1 & 45.5 & 27.2 & 45.8 & 57.1 \\
\rowcolor{mygray}
\textbf{\ourmethod-T++} & 301G & 45M & 3x & \textbf{50.5} & 71.9 & 55.7 & 35.0 & 54.3 & 65.3 & \textbf{45.1} & 69.2 & 48.7 & 26.7 & 48.5 & 64.0 \\
\hline
ConvNeXt-S~\cite{convnext} & 348G & 70M & 1x & 45.4 & 67.9 & 50.0 & - & - & - & 41.8 & 65.2 & 45.1 & - & - & - \\
RegionViT-B+~\cite{rgvit} & 307G & 93M & 1x & 45.4 & 68.4 & 49.6 & 28.5 & 48.7 & 60.2 & 41.6 & 65.2 & 44.8 & 21.6 & 44.4 & 60.5 \\ 
Swin-S~\cite{Swin} & 359G & 69M & 1x & 45.7 & 67.9 & 50.4 & 29.5 & 48.9 & 60.0 & 41.1 & 64.9 & 44.2 & 25.1 & 44.3 & 56.6 \\
ConvNeXt-B~\cite{convnext} & 486G & 108M & 1x & 47.0 & 69.4 & 51.7 & - & - & - & 42.7 & 66.3 & 46.0 & - & - & - \\
PVTv2-B5~\cite{PVTv2} & 557G & 102M & 1x & 47.4 & 68.6 & 51.9 & 28.8 & 51.0 & 63.1 & 42.5 & 65.7 & 46.0 & 22.4 & 45.8 & 61.1 \\
PVTv2-B4~\cite{PVTv2} & 500G & 82M & 1x & 47.5 & 68.7 & 52.0 & 30.1 & 50.9 & 62.9 & 42.7 & 66.1 & 46.1 & 23.3 & 45.6 & 62.0 \\
InternImage-S~\cite{DCNv3} & 340G & 69M & 1x & 47.8 & 69.8 & 52.8 & 30.4 & 51.8 & 62.7 & 43.3 & 67.1 & 46.7 & 23.1 & 46.8 & 62.5 \\
CSwin-S~\cite{cswin} & 342G & 54M & 1x & 47.9 & 70.1 & 52.6 & - & - & - & 43.2 & 67.1 & 46.2 & - & - & - \\
CSwin-B~\cite{cswin} & 526G & 97M & 1x & 48.7 & 70.4 & 53.9 & - & - & - & 43.9 & 67.8 & 47.3 & - & - & - \\
InternImage-B~\cite{DCNv3} & 501G & 115M & 1x & 48.8 & 70.9 & 54.0 & 31.9 & 52.4 & 63.1 & 44.0 & 67.8 & 47.4 & 24.3 & 47.2 & 62.5 \\
\rowcolor{mygray}
\ourmethod-S~\cite{Xia_2022_CVPR} & 378G & 69M & 1x & 47.1 & 69.9 & 51.5 & 30.5 & 50.1 & 62.1 & 42.5 & 66.7 & 45.4 & 25.5 & 45.8 & 58.5 \\
\rowcolor{mygray}
\textbf{\ourmethod-S++} & 430G & 74M & 1x & \textbf{49.8} & 71.9 & 54.6 & 33.8 & 53.9 & 64.4 & \textbf{44.5} & 68.7 & 48.2 & 25.0 & 48.0 & 63.3 \\
\hline
ConvNeXt-S~\cite{convnext} & 348G & 70M & 3x & 47.9 & 70.0 & 52.7 & - & - & - & 42.9 & 66.9 & 46.2 & - & - & - \\
RegionViT-B+~\cite{rgvit} & 307G & 93M & 3x & 48.3 & 70.1 & 52.8 & 31.8 & 51.1 & 64.3 & 43.5 & 67.1 & 47.0 & 25.1 & 46.4 & 62.1 \\
NAT-S~\cite{nat} & 330G & 70M & 3x & 48.4 & 69.8 & 53.2 & 31.9 & 52.1 & 62.2 & 43.2 & 66.9 & 46.4 & 23.3 & 46.6 & 61.8 \\
ConvNeXt-B~\cite{convnext} & 486G & 108M & 3x & 48.5 & 70.1 & 53.3 & - & - & - & 43.5 & 67.1 & 46.7 & - & - & - \\
Swin-S~\cite{Swin} & 359G & 69M & 3x & 48.5 & 70.2 & 53.5 & 33.4 & 52.1 & 63.3 & 43.3 & 67.3 & 46.6 & 28.1 & 46.7 & 58.6 \\
DiNAT-S~\cite{dinat} & 330G & 70M & 3x & 49.3 & 70.8 & 54.2 & 34.0 & 53.0 & 63.7 & 43.9 & 68.0 & 47.4 & 24.4 & 47.1 & 63.1 \\
InternImage-S~\cite{DCNv3} & 340G & 69M & 3x & 49.7 & 71.1 & 54.5 & 33.0 & 53.3 & 64.4 & 44.5 & 68.5 & 47.8 & 24.8 & 47.7 & 62.9 \\ 
CSwin-S~\cite{cswin} & 342G & 54M & 3x & 50.0 & 71.3 & 54.7 & - & - & - & 44.5 & 68.4 & 47.7 & - & - & - \\
InternImage-B~\cite{DCNv3} & 501G & 115M & 3x & 50.3 & 71.4 & 55.3 & 35.3 & 53.5 & 64.6 & 44.8 & 68.7 & 48.0 & 26.8 & 47.6 & 63.1 \\
CSwin-B~\cite{cswin} & 526G & 97M & 3x & 50.8 & 72.1 & 55.8 & - & - & - & 44.9 & 69.1 & 48.3 & - & - & - \\
\rowcolor{mygray}
\ourmethod-S~\cite{Xia_2022_CVPR} & 378G & 69M & 3x & 49.0 & 70.9 & 53.8 & 32.7 & 52.6 & 64.0 & 44.0 & 68.0 & 47.5 & 27.8 & 47.7 & 59.5 \\
\rowcolor{mygray}
\textbf{\ourmethod-S++} & 430G & 74M & 3x & \textbf{51.2} & 72.6 & 56.3 & 35.8 & 55.4 & 65.6 & \textbf{45.7} & 69.9 & 49.7 & 27.6 & 49.2 & 64.3 \\
\thickhline
\end{tabular}}}
\vspace{-0.6cm}
\end{center}
\end{table*}

\begin{table*}[t]
\begin{center}
\caption{Object detection and instance segmentation performance on MS-COCO dataset with Cascade Mask R-CNN~\cite{cmrcn} detector. This table shows the number of parameters, computational cost (FLOPs), mAP at different IoU thresholds, and different object sizes. The FLOPs are computed over the entire detector at the resolution of 1280$\times$800$\times$3. 
}\label{tab:det_cmrcn}
\vspace{-0.2in}
\setlength{\tabcolsep}{1.2mm}{
\renewcommand\arraystretch{1.2}
\resizebox{\linewidth}{!}{
\begin{tabular}{l|c|c|c|ccc|ccc|ccc|ccc}
\thickhline
\multicolumn{16}{c}{\textbf{Cascade Mask R-CNN Object Detection \& Instance Segmentation on MS-COCO}} \\
Method & FLOPs & \#Param & Schedule & AP$^b$ & AP$^b_\text{50}$ & AP$^b_\text{75}$ & AP$^b_s$ & AP$^b_m$ & AP$^b_l$ & AP$^m$ & AP$^m_\text{50}$ & AP$^m_\text{75}$ & AP$^m_s$ & AP$^m_m$ & AP$^m_l$ \\
\hline
Swin-T~\cite{Swin} & 745G & 86M & 1x & 48.1 & 67.1 & 52.2 & 30.4 & 51.5 & 63.1 & 41.7 & 64.4 & 45.0 & 24.0 & 45.2 & 56.9 \\
\rowcolor{mygray}
\ourmethod{}-T~\cite{Xia_2022_CVPR} & 750G & 86M & 1x & 49.1 & 68.2 & 52.9 & 31.2 & 52.4 & 65.1 & 42.5 & 65.4 & 45.8 & 25.2 & 45.9 & 58.6 \\ 
\rowcolor{mygray}
\textbf{\ourmethod{}-T++} & 771G & 81M & 1x & \textbf{52.2} & 70.9 & 56.6 & 33.9 & 56.2 & 68.1 & \textbf{45.0} & 68.1 & 48.9 & 25.1 & 48.5 & 63.4 \\
\hline
Swin-T~\cite{Swin} & 745G & 86M & 3x & 50.4 & 69.2 & 54.7 & 33.8 & 54.1 & 65.2 & 43.7 & 66.6 & 47.3 & 27.3 & 47.5 & 59.0 \\
ConvNeXt-T~\cite{convnext} & 741G & 86M & 3x & 50.4 & 69.1 & 54.8 & 33.9 & 54.4 & 65.1 & 43.7 & 66.5 
 & 47.3 & 24.2 & 47.1 & 62.0 \\
NAT-T~\cite{nat} & 737G & 85M & 3x & 51.4 & 70.0 & 55.9 & 33.8 & 55.2 & 66.2 & 44.5 & 67.6 & 47.9 & 23.7 & 47.8 & 63.4 \\
DiNAT-T~\cite{dinat} & 737G & 85M & 3x & 52.2 & 71.0 & 56.9 & 36.1 & 55.9 & 67.9 & 45.2 & 68.3 & 48.9 & 26.4 & 48.4 & 64.5 \\
HorNet-T~\cite{hornet} & 728G & 80M & 3x & 52.4 & 71.6 & 56.8 & 36.4 & 56.8 & 67.1 & 45.6 & 69.1 & 49.6 & 26.5 & 49.2 & 64.3 \\
CSWin-T~\cite{cswin} & 757G & 80M & 3x & 52.5 & 71.5 & 57.1 & - & - & - & 45.3 & 68.8 & 48.9 & - & - & - \\
\rowcolor{mygray}
\ourmethod-T~\cite{Xia_2022_CVPR} & 750G & 86M & 3x & 51.3 & 70.1 & 55.8 & 34.1 & 54.6 & 66.9 & 44.5 & 67.5 & 48.1 & 27.9 & 47.9 & 60.3 \\
\rowcolor{mygray}
\textbf{\ourmethod{}-T++} & 771G & 81M & 3x & \textbf{53.0} & 71.6 & 57.7 & 37.1 & 56.6 & 68.6 & \textbf{46.0} & 69.3 & 50.1 & 26.7 & 49.3 & 64.3 \\
\hline
Swin-S~\cite{Swin} & 838G & 107M & 3x & 51.9 & 70.7 & 56.3 & 35.2 & 55.7 & 67.7 & 45.0 & 68.2 & 48.8 & 28.8 & 48.7 & 60.6 \\
ConvNeXt-S~\cite{convnext} & 827G & 108M & 3x & 51.9 & 70.8 & 56.5 & 35.4 & 55.7 & 67.3 & 45.0 & 68.4 & 49.1 & 25.2 & 48.4 & 64.0 \\
NAT-S~\cite{nat} & 809G & 108M & 3x & 51.9 & 70.4 & 56.2 & 35.7 & 55.9 & 66.8 & 44.9 & 68.2 & 48.6 & 25.8 & 48.5 & 63.2 \\
DiNAT-S~\cite{dinat} & 809G & 108M & 3x & 52.9 & 71.8 & 57.5 & 37.6 & 56.8 & 68.3 & 45.8 & 69.3 & 49.8 & 26.5 & 49.4 & 64.7 \\
HorNet-S~\cite{hornet} & 827G & 108M & 3x & 53.3 & 72.3 & 57.8 & 36.7 & 57.8 & 69.0 & 46.3 & 69.9 & 50.4 & 26.8 & 50.2 & 65.1 \\
CSWin-S~\cite{cswin} & 820G & 92M & 3x & 53.7 & 72.2 & 58.4 & - & - & - & 46.4 & 69.6 & 50.6  & - & - & - \\
\rowcolor{mygray}
\ourmethod-S~\cite{Xia_2022_CVPR} & 857G & 107M & 3x & 52.7 & 71.7 & 57.2 & 37.3 & 56.3 & 68.0 & 45.5 & 69.1 & 49.3 & 30.2 & 49.2 & 60.9 \\
\rowcolor{mygray}
\textbf{\ourmethod{}-S++} & 895G & 110M & 3x & \textbf{54.2} & 72.7 & 58.9 & 38.0 & 58.3 & 69.7 & \textbf{46.9} & 70.1 & 51.3 & 28.3 & 50.3 & 65.8 \\
\hline
Swin-B~\cite{Swin} & 982G & 145M & 3x & 51.9 & 70.5 & 56.4 & 35.4 & 55.2 & 67.4 & 45.0 & 68.1 & 48.9 & 28.9 & 48.3 & 60.4 \\
NAT-B~\cite{nat} & 931G & 147M & 3x & 52.2 & 70.9 & 56.8 & 35.2 & 55.9 & 67.2 & 45.1 & 68.3 & 49.1 & 25.5 & 48.4 & 64.0 \\
ConvNeXt-B~\cite{convnext} & 964G & 146M & 3x & 52.7 & 71.4 & 57.2 & 36.5 & 56.6 & 68.2 & 45.6 & 68.9 & 49.6 & 26.0 & 49.0 & 64.6 \\
DiNAT-B~\cite{dinat} & 931G & 147M & 3x & 53.4 & 72.0 & 58.2 & 37.4 & 57.5 & 68.8 & 46.2 & 69.7 & 50.2 & 26.4 & 49.7 & 65.1 \\
CSWin-B~\cite{cswin} & 1004G & 135M & 3x & 53.9 & 72.6 & 58.5 & - & - & - & 46.4 & 70.0 & 50.4 & - & - & - \\
HorNet-B~\cite{hornet} & 965G & 146M & 3x & 54.0 & 72.8 & 58.7 & 37.8 & 58.1 & 69.7 & 46.9 & 70.2 & 51.1 & 27.4 & 50.4 & 65.7 \\
\rowcolor{mygray}
\ourmethod-B~\cite{Xia_2022_CVPR} & 1003G & 145M & 3x & 53.0 & 71.9 & 57.6 & 36.0 & 56.8 & 69.1 & 45.8 & 69.3 & 49.5 & 29.2 & 49.5 & 61.9 \\
\rowcolor{mygray}
\textbf{\ourmethod{}-B++} & 1059G & 151M & 3x & \textbf{54.5} & 73.0 & 59.4 & 38.5 & 58.4 & 69.8 & \textbf{47.0} & 70.5 & 51.4 & 27.9 & 50.3 & 65.8 \\
\thickhline
\end{tabular}}}
\vspace{-0.3cm}
\end{center}
\end{table*}

\noindent
\textbf{Results.} We report the image classification results of~\ourmethod{}++ on ImageNet-1K~\cite{in1k} in Table~\ref{tab:in1k}. We compare our models with recent state-of-the-art ViTs and CNNs at three different model scales regarding FLOPs and parameters. Compared to other methods,~\ourmethod{}++ achieves comparable or even superior accuracy of the ImageNet-1K validation set at similar computational complexity or model size. For instance,~\ourmethod{}-T++ achieves 83.9$\%$ Top-1 accuracy with only 4.3G FLOPs, surpassing previous strong SOTA models~\cite{hornet,DCNv3,biformer}. 

As the models scale up,~\ourmethod{}-S/B++ also outperform or compete with these methods. For models at the \textit{Small} scale, ~\ourmethod{}-S++ outperforms Focal Transformer~\cite{focal}, NAT~\cite{nat}, and DiNAT~\cite{dinat} with significant gains in accuracy from 0.8 to 1.0, obtaining 84.6\% Top-1 accuracy. Even compared to some more recent leading ViTs/CNNs such as HorNet~\cite{hornet}, InternImage~\cite{DCNv3}, and BiFormer~\cite{biformer}, ~\ourmethod{}++ also shows competitive results. The superiority of the proposed model is tenable on the \textit{Base} scale models, where~\ourmethod{}-B++ achieves 84.9\% and 85.9\% classification accuracy at the input resolution of 224$\times{}$224 and 384$\times{}$384, respectively.

\subsection{Object Detection and Instance Segmentation}\label{sec:e_m_det}

\noindent
\textbf{Settings.} MS-COCO~\cite{coco} dataset has about 118K training images and 5K validation images, which is a widely accepted benchmark for object detection and instance segmentation. To validate our proposed models on downstream vision tasks, we use~\ourmethod{}++ as a backbone network and incorporate it into the detectors to extract visual features from the images containing various objects and instances. ~\ourmethod{}++ is employed in three widely used detectors: RetinaNet~\cite{rtn}, Mask R-CNN~\cite{mrcn}, and Cascade Mask R-CNN~\cite{cmrcn}, and initialized from the ImageNet pretrained model weights with 300 epochs. Following common practice~\cite{Swin,PVT,convnext}, we train object detection and instance segmentation models with~\ourmethod{}++ for 12 (1x schedule) or 36 (3x schedule) epochs. 

In practice, we train the models on 32 GPUs with a linearly scaled learning rate of $4\times{}10^{-3}$ and batch size of 64. Since different training durations usually cause different convergence, we adjust the stochastic depth factor for each model on each scale and schedule to prevent overfitting. For RetinaNet~\cite{rtn}, this factor is set to 0.0 (1x), 0.2 (3x) for \ourmethod{}-T++, and 0.1 (1x), 0.5 (3x) for \ourmethod{}-S++. For Mask R-CNN~\cite{mrcn}, the drop path rates are 0.0 (1x), 0.3 (3x) for \ourmethod{}-T++, and 0.1 (1x), 0.5 (3x) for \ourmethod{}-S++. For Cascade Mask-RCNN~\cite{cmrcn}, the rates are 0.0 (1x), 0.1 (3x) for \ourmethod{}-T++, 0.5 (3x) for \ourmethod{}-S++, and 0.8 (3x) for \ourmethod{}-B++.

\noindent
\textbf{Results.} Table~\ref{tab:det_retinanet} shows the object detection results of ~\ourmethod{}-T/S++ on RetinaNet~\cite{rtn} in the 1x and 3x schedules.
The \textit{tiny} variant achieves 46.8 mAP on 1x schedule and 49.2 mAP on 3x schedule, surpassing other methods by at least 1.5 mAP, and the \textit{small} variant with respective 48.3 mAP (1x) and 50.2 mAP (3x) exhibits significant improvements over other baselines with similar parameters and FLOPs.

As for instance segmentation, we combine the \textit{Tiny}, \textit{Small} scale variants of ~\ourmethod{}++ on Mask R-CNN~\cite{mrcn} in the 1x and 3x schedules. As shown in Table~\ref{tab:det_mrcn}, \ourmethod{}-T++ achieves consistent improvements on several SOTA models~\cite{cswin, DCNv3} by up to 1.5 bbox mAP and 1.4 mask mAP. \ourmethod{}-S++ even outperforms some \textit{Base} variants of the baselines with at least 0.4 bbox mAP and 0.5 mask mAP. Similar performance gains are also observed in Cascade Mask R-CNN~\cite{cmrcn} framework. Table~\ref{tab:det_cmrcn} reports that ~\ourmethod{}++ reaches a new level of precision with 53.0 bbox mAP, 46.0 mask mAP of \textit{Tiny}, 54.2, 46.9 of \textit{Small}, 54.6, 47.2 of \textit{Base}, respectively, which show promising margins over other strong ViT/CNN models. 

\begin{table}[t]
\newcommand{\tabincell}[2]{\begin{tabular}{@{}#1@{}}#2\end{tabular}}
\begin{center}
\caption{Semantic segmentation performance on ADE20K dataset with SemanticFPN~\cite{semfpn} and UperNet~\cite{upernet} segmentors. The FLOPs are computed over the whole segmentor at the resolution of 2048$\times$512$\times$3. $\ddagger$ means testing model with multi scale and flip augmentations.}
\label{tab:seg}
\vspace{-0.2in}
\setlength{\tabcolsep}{1.0mm}{
\renewcommand\arraystretch{1.25}
\resizebox{\linewidth}{!}{
\begin{tabular}{c|l|cc|cc}
\thickhline
\multicolumn{6}{c}{\textbf{Semantic Segmentation on ADE20K}} \\
Method & Backbone & FLOPs & \#Param & mIoU & mIoU$^\ddagger$ \\
\hline
\multirow{12}{*}{\tabincell{c}{Seman-\\ticFPN}} & PVT-S~\cite{PVT} & 225G & 28M & 42.0 & 42.0 \\
& PVTv2-B2~\cite{PVTv2} & 230G & 29M & 45.2 & - \\
& RegionViT-S+~\cite{rgvit} & 236G & 35M & 45.3 & - \\
& \cellcolor{mygray}\ourmethod-T~\cite{Xia_2022_CVPR} \cellcolor{mygray}& \cellcolor{mygray}198G & \cellcolor{mygray}32M & \cellcolor{mygray}42.6 & \cellcolor{mygray}44.2 \\
& \cellcolor{mygray}\textbf{\ourmethod-T++} & \cellcolor{mygray}212G & \cellcolor{mygray}28M & \cellcolor{mygray}\textbf{48.4} & \cellcolor{mygray}\textbf{48.8} \\
\cline{2-6}
& PVT-M~\cite{PVT} & 315G & 48M & 42.9 & 43.3 \\
& RegionViT-B+~\cite{rgvit} & 459G & 77M & 47.5 & - \\
& PVTv2-B4~\cite{PVTv2} & 427G & 66M & 47.9 & - \\
& \cellcolor{mygray}\ourmethod-S~\cite{Xia_2022_CVPR} & \cellcolor{mygray}320G & \cellcolor{mygray}53M & \cellcolor{mygray}46.1 & \cellcolor{mygray}48.5 \\
& \cellcolor{mygray}\textbf{\ourmethod-S++} & \cellcolor{mygray}339G & \cellcolor{mygray}57M & \cellcolor{mygray}\textbf{49.9} & \cellcolor{mygray}\textbf{50.7} \\
\cline{2-6}
& PVT-L~\cite{PVT} & 420G & 65M & 43.5 & 43.9 \\
& PVTv2-B5~\cite{PVTv2} & 486G & 86M & 48.7 & - \\
& \cellcolor{mygray}\ourmethod-B~\cite{Xia_2022_CVPR} & \cellcolor{mygray}481G & \cellcolor{mygray}92M & \cellcolor{mygray}47.0 & \cellcolor{mygray}49.0 \\
& \cellcolor{mygray}\textbf{\ourmethod-B++} & \cellcolor{mygray}508G & \cellcolor{mygray}97M & \cellcolor{mygray}\textbf{50.4} & \cellcolor{mygray}\textbf{51.1} \\
\hline
\multirow{27}{*}{UperNet} & Swin-T~\cite{Swin} & 945G & 60M & 44.5 & 45.8 \\
& ConvNeXt-T~\cite{convnext} & 939G & 60M & 46.1 & 46.7 \\
& Focal-T~\cite{focal} & 998G & 62M & 45.8 & 47.0 \\
& InternImage-T~\cite{DCNv3} & 944G & 59M & 47.9 & 48.1 \\
& NAT-T~\cite{nat} & 934G & 58M & 47.1 & 48.4 \\
& DiNAT-T~\cite{dinat} & 934G & 58M & 47.8 & 48.8 \\
& HorNet-T~\cite{hornet} & 924G & 55M & 49.2 & 49.3 \\
& \cellcolor{mygray}\ourmethod-T~\cite{Xia_2022_CVPR} & \cellcolor{mygray}957G & \cellcolor{mygray}60M & \cellcolor{mygray}45.5 & \cellcolor{mygray}46.4 \\
& \cellcolor{mygray}\textbf{\ourmethod-T++} & \cellcolor{mygray}969G & \cellcolor{mygray}54M & \cellcolor{mygray}\textbf{49.4} & \cellcolor{mygray}\textbf{50.3}\\
\cline{2-6}
& Swin-S~\cite{Swin} & 1038G & 81M & 47.6 & 49.5 \\
& NAT-S~\cite{nat} & 1010G & 82M & 48.0 & 49.5 \\
& ConvNeXt-S~\cite{convnext} & 1027G & 82M & 48.7 & 49.6 \\
& DiNAT-S~\cite{dinat} & 1010G & 82M & 48.9 & 49.9 \\
& Focal-S~\cite{focal} & 1130G & 85M & 48.0 & 50.0 \\
& HorNet-S~\cite{hornet} & 1027G & 85M & 50.0 & 50.5 \\
& InternImage-S~\cite{DCNv3} & 1017G & 80M & 50.1 & 50.9 \\
& \cellcolor{mygray}\ourmethod-S~\cite{Xia_2022_CVPR} & \cellcolor{mygray}1079G & \cellcolor{mygray}81M & \cellcolor{mygray}48.3 & \cellcolor{mygray}49.8  \\
& \cellcolor{mygray}\textbf{\ourmethod-S++} & \cellcolor{mygray}1098G & \cellcolor{mygray}85M & \cellcolor{mygray}\textbf{50.5} & \cellcolor{mygray}\textbf{51.2}\\
\cline{2-6}
& Swin-B~\cite{Swin} & 1188G & 121M	& 48.1 & 49.7 \\
& NAT-B~\cite{nat} & 1137G & 123M & 48.5 & 49.7 \\
& ConvNeXt-B~\cite{convnext} & 1170G & 122M & 49.1 & 49.9 \\
& DiNAT-B~\cite{dinat} & 1137G & 123M & 49.6 & 50.4 \\
& Focal-B~\cite{focal} & 1354G & 126M & 49.0 & 50.5 \\
& HorNet-B~\cite{hornet} & 1171G & 126M & 50.5 & 50.9 \\
& InternImage-B~\cite{DCNv3} & 1185G & 128M & 50.8 & 51.3 \\
& \cellcolor{mygray}\ourmethod-B~\cite{Xia_2022_CVPR} & \cellcolor{mygray}1212G & \cellcolor{mygray}121M & \cellcolor{mygray}49.4 & \cellcolor{mygray}50.6 \\
& \cellcolor{mygray}\textbf{\ourmethod-B++} & \cellcolor{mygray}1268G & \cellcolor{mygray}127M & \cellcolor{mygray}\textbf{51.0} & \cellcolor{mygray}\textbf{51.5}\\
\thickhline
\end{tabular}}}
\vspace{-0.4cm}
\end{center}
\end{table}

\subsection{Semantic Segmentation}\label{sec:e_m_seg}

\noindent
\textbf{Settings.} ADE20K~\cite{ade20k} is a popular dataset for semantic segmentation with 20K training examples and 2K validation examples. We employ our \ourmethod{} on two widely adopted segmentation models, SemanticFPN~\cite{semfpn} and UperNet~\cite{upernet}. Both frameworks share an encoder-decoder structure where we employ variants of \ourmethod{}++ as encoder networks. We follow the recipes in Swin Transformer~\cite{Swin} and PVT~\cite{PVT} to initialize the encoder from ImageNet pretrained weights, then train 160K iterations for UperNet and 80K for SemanticFPN. We train \ourmethod{}++ on the top of SemanticFPN~\cite{semfpn} with 0.4, 0.4, 0.7 drop path rates for \textit{Tiny}, \textit{Small}, and \textit{Base} variants. And for UperNet~\cite{upernet}, these factors are 0.3, 0.5, 0.7 for \ourmethod{}-T/S/B++, respectively.

\noindent
\textbf{Results.} Table~\ref{tab:seg} presents the results of different variants of \ourmethod{}++ with two segmentation frameworks on ADE20K. The first part lists the results with SemanticFPN, from which \ourmethod{}-T/S/B++ achieves 48.4, 49.9, and 50.4 in single-scale mIoU, 51.1 in multi-scale mIoU, respectively, which leads to sharp performance boosts over other methods. The similar superiority of \ourmethod{}++ holds for UperNet, as shown in the second part. Taking ConvNeXt~\cite{hornet} as an example, \ourmethod{}++ has a consistent improvement over HorNet on each of the three model scales, with +3.6, +2.6 and +1.6 in multi-scale mIoU, respectively. In addition, \ourmethod{}++ outperforms many baselines and performs similar to SOTA methods, such as HorNet~\cite{hornet} and InternImage~\cite{DCNv3}.


\subsection{Ablation Study}\label{sec:e_abl}

To better understand our model, we study the key components in the proposed DMHA module and some important design choices of \ourmethod{} and \ourmethod{}++ individually, including offset learning, deformable positional encoding, local modeling, and macro designs. We first sketch a roadmap from the original \ourmethod~\cite{Xia_2022_CVPR} to \ourmethod++ with \textit{Tiny} variant, displaying the steps toward a more powerful model with clean and effective modifications. We then select a version of \ourmethod-T++ without augmented convolutions, denoted as \ourmethod-T+, to ablate the design choices. At the end of this section, we provide a detailed analysis on the augmented convolutions to reveal how \ourmethod-T{}+ is transformed into \ourmethod-T{}++.

\noindent
\textbf{Roadmap from \ourmethod{} to \ourmethod++.} 
As shown in Figure~\ref{fig:roadmap}, we begin by removing the designs in the original \ourmethod{}-T, which require too many extra hyper-parameters to build a concise architecture. Reducing the number of deformed keys and values from $14^2$ to $7^2$ in the 3$^\text{rd}$ stage makes it unified to incorporate DMHA modules in other stages with limited computational complexity, with $7^2$ sampling points in all four stages. Another dispensable design is the restriction of the range of sampling offsets. The ablation in~\cite{Xia_2022_CVPR} makes it clear that a wide range from 1.0 to 10.0 of this restriction factor works well in many scenarios. Thus we remove it and just clip the sampling coordinates to ensure that these locations lie in the image. The above modifications bring about a -0.3 decline in accuracy but lead to a cleaner design. 

To achieve better performance, we adopt several amendments in macro design. The deeper and narrower structure has been shown to be effective in many previous studies~\cite{regnet,efficientnet}, and therefore we increase the number of blocks from 6 to 18 in stage 3 of \ourmethod{} while decreasing the dimensions of the basic model from 96 to 64 to alleviate the growth of complexity. In addition to depth and width, Neighborhood Attention~\cite{nat} provides an efficient implementation of local attention with higher accuracy-speed performance than sliding window ones~\cite{san,sasa}. We replace Window Attention~\cite{Swin} with Neighborhood Attention as the local attention module in \ourmethod{}++. In addition, we extend the DMHA to the first two stages to build a full deformable attention model and further add two blocks in stage 2, leading to an accuracy of 83.0 with 4.05 GFLOPs. We denote this version as \textbf{\ourmethod{}+} to ablate the design choices.

To further push the limit, we enhance \ourmethod{}+ with three convolutional modules introduced in Sec.~\ref{sec:m_conv} and end up with the final \ourmethod{}++ with 83.9 accuracy and 4.29 GFLOPs, which achieves state-of-the-art performance. The convolutional components are analyzed as follows.

\noindent
\textbf{Core components and designs in DMHA.}
We study the core components and designs in the proposed DMHA that include learning offset and exploiting geometric information to demonstrate the effectiveness of \ourmethod{}. We summarize these experiments in Table~\ref{tab:abl_off}, in which the first part shows the cases of disabling offsets, and the second part shows the influence of different types of position encoding methods. \ourmethod{}-T+ is shown in the last row, with 83.0 ImageNet-1K Top-1 accuracy and 46.2, 41.8 bbox and mask mAP on MS-COCO and 47.9 mIoU on ADE20K. To study the effect of offset learning, we come up with two variants of \ourmethod{}+ without learning any sampling offsets for deformed keys and values by replacing the DMHA module with the SRA module in PVT~\cite{PVT} or simply zeroing out the offsets to sample features at the reference points. These two variants are shown in the first two rows, from which the sharp drops of up to 1.1 accuracy, 0.6 mAP and 1.4 mIoU demonstrate the importance of learning offsets and focusing on important image parts in the proposed DMHA module.

\begin{table}[t]
\newcommand{\tabincell}[2]{\begin{tabular}{@{}#1@{}}#2\end{tabular}}
\begin{center}
\caption{Ablation study on the core components and designs in DMHA. \textbf{P} represents replacing DMHA with SRA attention proposed in \cite{PVT}, while \textbf{D} denotes the proposed DMHA attention module. \ding{51} in offsets means using learnable offsets to guide deformable feature sampling whereas \ding{55} means not. The FLOPs are computed at the resolution of $224^2$ of the IN-1K classification model. Object detection is done with 1x Mask R-CNN and semantic segmentation is done with 160K UperNet.}
\label{tab:abl_off}
\vspace{-0.2in}
\setlength{\tabcolsep}{0.3mm}{
\renewcommand\arraystretch{1.3}
\resizebox{\linewidth}{!}{
\begin{tabular}{ccc|cc|cccc}
\thickhline
\multicolumn{3}{c|}{Components} & \multicolumn{2}{c|}{Backbone} & IN-1K & \multicolumn{2}{c}{MS-COCO} & ADE20K \\
Attn. & Off. & Pos. Enc. & FLOPs & \#Param & Acc@1 & AP$^b$ & AP$^m$ & mIoU \\
\hline
\textbf{\textsf{P}} & \ding{55} & \ding{55} & 4.20G & 25.8M & 81.9 & 45.6 & 41.3 & 47.2 \\
\textbf{\textsf{D}} & \ding{55} & \ding{55} & 4.05G & 22.6M & 82.4 & 45.7 & 41.4 & 46.5 \\
\hline
\textbf{\textsf{D}} & \ding{51} & \ding{55} & 4.05G & 22.6M & 82.8 & 45.8 & 41.6 & 47.5 \\
\textbf{\textsf{D}} & \ding{51} & Fixed & 4.05G & 24.0M & 82.8 & 45.8 & 41.5 & 47.8 \\
\textbf{\textsf{D}} & \ding{51} & LogCPB~\cite{Swinv2} & 4.13G & 22.7M & 82.5 & \textbf{46.3} & 41.6 & 47.8 \\
\hline
\rowcolor{mygray}
\textbf{\textsf{D}} & \ding{51} & DeformRPB & 4.05G & 22.8M & \textbf{83.0} & 46.2 & \textbf{41.8} & \textbf{47.9} \\
\thickhline
\end{tabular}}}
\vspace{0.2cm}
\caption{Ablation study on different types of local modeling operators. These modules play roles of \textbf{Local Attention} in Figure~\ref{fig:fig_model_arch}. FLOPs and parameters are computed at the resolution of $224^2$ of image classification model. Our proposed \textbf{\ourmethod{}+} is marked with gray line.}
\label{tab:abl_local}
\vspace{-0.3cm}
\setlength{\tabcolsep}{0.3mm}{
\renewcommand\arraystretch{1.3}
\resizebox{\linewidth}{!}{
\begin{tabular}{c|cc|cccc}
\thickhline
\multirow{2}{*}{Local Operators} & \multicolumn{2}{c|}{Backbone} & IN-1K & \multicolumn{2}{c}{MS-COCO} & ADE20K \\
& FLOPs & \#Param & Acc@1 & AP$^b$ & AP$^m$ & mIoU \\
\hline
Window Attn.~\cite{Swin} & 4.05G & 22.8M & 82.4 & 44.9 & 41.0 & 46.8 \\
Slide Attn.~\cite{slide} & 3.57G & 20.9M & 82.5 & 45.9 & 41.6 & 47.7 \\
Depth-wise Conv.~\cite{convnext} & 3.39G & 20.4M & 82.3 & 46.1 & \textbf{41.8} & 47.7 \\
\hline
\rowcolor{mygray}
Neighborhood Attn.~\cite{nat} & 4.05G & 22.8M & \textbf{83.0} & \textbf{46.2} & \textbf{41.8} & \textbf{47.9} \\
\thickhline
\end{tabular}}}
\vspace{-0.6cm}
\end{center}
\end{table}

To study the most suitable way of exploiting geometric information for DMHA, we compare the performance of different position encoding methods in the third to fifth rows in Table~\ref{tab:abl_off}. Enabling offset learning without using any position encoding brings about +0.4$\%$ accuracy, +0.1 mAP and +0.3 mIoU on different tasks. However, it makes hardly any difference to simply involve a fixed position bias in the deformable attention since there are no improvements on image classification and object detection tasks, except for more parameters. We also try a light version of the LogCPB proposed in~\cite{Swinv2}, by reducing the hidden dimensions of the MLP from 512 to 32 in order to maintain a feasible computation complexity and memory footprint. LogCPB exhibits good performance on object detection tasks with higher input resolutions, but image classification accuracy has a degradation of 0.3$\%$. In addition, as the deformable attention works globally on the feature maps, LogCPB has larger FLOPs than other and scales poorly to high-resolution inputs. The last row reports the results of our proposed deformable relative positional bias, showing strong overall performance on various tasks, indicating that DeformRPB is the best suitable approach to exploiting geometric information in the scheme of deformable attention.

\begin{table}[t]
\newcommand{\tabincell}[2]{\begin{tabular}{@{}#1@{}}#2\end{tabular}}
\begin{center}
\caption{Ablation study on applying DMHA on different stages. Configuration \textbf{ND} denotes this stage is composed of successive Neighborhood Attention and DMHA blocks, as \textbf{NN} and \textbf{DD} denote the stage is made up of the same blocks of Neighborhood Attention and DMHA, respectively. }
\label{tab:abl_ds}
\vspace{-0.2in}
\setlength{\tabcolsep}{0.5mm}{
\renewcommand\arraystretch{1.4}
\resizebox{\linewidth}{!}{
\begin{tabular}{p{0.6cm}<{\centering}p{0.6cm}<{\centering}p{0.6cm}<{\centering}p{0.6cm}<{\centering}|cc|cccc}
\thickhline
\multicolumn{4}{c|}{Stage Configurations} & \multicolumn{2}{c|}{Backbone} & IN-1K & \multicolumn{2}{c}{MS-COCO} & ADE20K \\
1 & 2 & 3 & 4 & FLOPs & \#Param & Acc@1 & AP$^b$ & AP$^m$ & mIoU \\
\hline
\textbf{\textsf{NN}} & \textbf{\textsf{NN}} & \textbf{\textsf{NN}} & \textbf{\textsf{NN}} & 4.29G & 22.7M & 81.4 & 43.9 & 39.7 & 45.3 \\
\textbf{\textsf{NN}} & \textbf{\textsf{NN}} & \textbf{\textsf{NN}} & \textbf{\textsf{ND}} & 4.29G & 22.7M & 81.9 & 45.0 & 40.9 & 45.5 \\
\textbf{\textsf{NN}} & \textbf{\textsf{NN}} & \textbf{\textsf{NN}} & \textbf{\textsf{DD}} & 4.29G & 22.7M & 82.3 & 45.2 & 41.2 & 46.7 \\
\textbf{\textsf{NN}} & \textbf{\textsf{NN}} & \textbf{\textsf{ND}} & \textbf{\textsf{DD}} & 4.21G & 22.7M & 82.8 & \textbf{46.4} & \textbf{41.8} & 47.5 \\
\textbf{\textsf{NN}} & \textbf{\textsf{ND}} & \textbf{\textsf{ND}} & \textbf{\textsf{DD}} & 4.08G & 22.7M & 83.0 & 46.3 & 41.7 & 47.7 \\
\hline
\rowcolor{mygray}
\textbf{\textsf{ND}} & \textbf{\textsf{ND}} & \textbf{\textsf{ND}} & \textbf{\textsf{DD}} & 4.05G & 22.8M & \textbf{83.0} & 46.2 & \textbf{41.8} & \textbf{47.9} \\
\thickhline
\end{tabular}}}
\vspace{0.5cm}
\caption{Ablation study on convolutional enhancements. ConvFFN means inserting a residual depth-wise convolution layer in MLP blocks, ConvStem means using overlapped patch embedding and downsampling modules between stages, LPU~\cite{cmt} means adding a residual depth-wise convolution layer before each attention block.}\label{tab:abl_conv}
\vspace{-0.3cm}
\setlength{\tabcolsep}{0.5mm}{
\renewcommand\arraystretch{1.3}
\resizebox{\linewidth}{!}{
\begin{tabular}{c|cc|cccc}
\thickhline
Added Convolution & \multicolumn{2}{c|}{Backbone} & IN-1K & \multicolumn{2}{c}{MS-COCO} & ADE20K \\
Modules & FLOPs & \#Param & Acc@1 & AP$^b$ & AP$^m$ & mIoU \\
\hline
None (\ourmethod{}-T+) & 4.05G & 22.8M & 83.0 & 46.2 & 41.8 & 47.9 \\
+ ConvFFN & 4.12G & 23.0M & 83.4 & 47.5 & 42.6 & 48.6 \\
+ ConvStem & 4.27G & 23.9M & 83.6 & 48.2 & 43.2 & 49.0 \\
\hline
\rowcolor{mygray}
+ LPU (\ourmethod{}-T++) & 4.29G & 24.0M & \textbf{83.9} & \textbf{48.7} & \textbf{43.7} & \textbf{49.4} \\
\thickhline
\end{tabular}}}
\vspace{-0.6cm}
\end{center}
\end{table}

\noindent
\textbf{Locality modeling operators.} 
We next investigate the combination of the proposed DMHA module with different operations to model the locality in the images. We replace the Neighborhood Attenion~\cite{nat} module with three other kinds of local operators: non-overlapped window attention in Swin Transformer~\cite{Swin}, 7$\times$7 convolution in ConvNeXt~\cite{convnext}, and an novel local attention named Slide Attention that leverages depth-wise convolution~\cite{slide} to achieve efficient sliding window attention. These modules play the role of \textbf{Local Attention} in Figure~\ref{fig:fig_model_arch}, whose performances on various vision tasks are listed in Table~\ref{tab:abl_local}. When using non-overlapped window attention, the model only gets 82.4$\%$ ImageNet accuracy, even falling behind the one with 82.6$\%$ in Figure~\ref{fig:roadmap} whose first two stages are identical to Swin Transformer. In addition, the window attention variant also behaves poorly on downstream tasks by -0.2 descent in object detection and semantic segmentation. The novel Slide Attention works much better than window attention with thorough improvements. Even though this variant has fewer parameters and less FLOPs, it has biquadratic memory footprints \textit{w.r.t.} the kernel size, making it viable only in 3$\times$3 kernels. Other than local attention methods, depth-wise convolutions can model locality efficiently, and many prior researchers have taken a step forward in modernizing CNNs with large kernels~\cite{convnext,replknet,slak}. We choose the carefully designed 7$\times$7 depth-wise convolution with MLP module in ConvNeXt to see its performance. The convolutional operator achieves similar results on MS-COCO object detection and ADE20K semantic segmentation with much fewer FLOPs, whereas it fails to close the gap in ImageNet classification. Therefore, we select Neighborhood Attention as the local attention module in \ourmethod{}++. These results also imply the versatility of \ourmethod{} with various local operators.

\noindent
\textbf{DMHA at different stages.}
We explore this design choice by gradually replacing Neighborhood Attention blocks with DMHA blocks stage by stage, as shown in Table~\ref{tab:abl_ds}. In the first row, all stages of the model are composed of Neighborhood Attention, which resembles NAT-T~\cite{nat} in the block configurations. This version only has 81.4\% accuracy in image classification, 43.9 mAP in object detection, and 45.3 mIoU in semantic segmentation. We begin by replacing one Neighborhood Attention block in the 4$^\text{th}$ stage with the proposed DMHA module, which immediately brings +0.5, +1.1, +1.2 and +0.2 gains on four metrics. Next we equip the whole 4$^\text{th}$ stage with DMHA, following as additional +0.4, +0.2, +0.3, +1.2 improvements. Keeping the procedure of DMHA replacements in the 3$^\text{rd}$ stage, more increments are obtained in these tasks. These sharp increments demonstrate the effectiveness of our proposed deformable attention. Although the increasing model performance as incorporating DMHA in more early stages is prone to saturate, we choose the final version with all four stages with deformable attention as \ourmethod{}-T+ to construct as simple a model as possible.

\noindent
\textbf{Convolutional enhancements.} Finally, we analyze how the convolutional enhancements in Sec.~\ref{sec:m_conv} can further boost the model performance and examine the their contribution. Table~\ref{tab:abl_conv} shows continuous improvements are made using ConvFFN, overlapped patch embeddings and downsamplings (+ConvStem), and local perception units (+LPU), step by step. We end up with \ourmethod{}++ with 83.9\% ImageNet-1K accuracy, 48.7 MS-COCO mAP, and 49.4 ADE20K mIoU, which is competitive with SOTA ViT/CNNs.

\subsection{\ourmethod{} and Deformable DETR}\label{sec:ddetr}

In this section, we compare two different types of deformable attention in \ourmethod{} and in Deformable-DETR~\cite{DeformableDETR}, which is a direct adaptation of deformable convolution~\cite{DCNv1}. We discuss the differences as follows:

\noindent
\textbf{The role of deformable attention.} In \ourmethod{}, deformable attention works in the backbone while serving as an efficient detection head in Deformable DETR to improve the convergence rate of the original DETR~\cite{detr}.

\noindent
\textbf{Attention mechanism.}
The $m$-th head of query $q$ in single scale attention of Deformable DETR is formulated as
\begin{equation}
z^{(m)}_q=\sum_{k=1}^{K}A^{(m)}_{qk}W_v\phi(x;p_q+\Delta{}p^{(m)}_{qk}),
\end{equation}
where $K$ keys are sampled from the input features, mapped by $W_v$ and then aggregated by attention weights $A^{(m)}_{qk}$. Compared to our Deformable Multi-Head Attention (DMHA, Eq.~\eqref{eq:dmha}), these attention weights are learned from $z_q$ by a linear projection, \textit{i.e.} $A^{(m)}_{qk}=\sigma(W_\text{att}x)$, where $W_\text{att}\!\in\!\mathbb{R}^{C\times{}MK}$ is a weight matrix to predict the weights for each key head by head, after which a softmax function $\sigma$ is applied to normalize the attention score of $K$ keys to $[0,1]$. In fact, the attention weights are directly predicted by queries instead of comparing the similarities between queries and keys in the dot product manner. If we change the $\sigma$ function to a sigmoid, it will be a variant of a modulated deformable convolution~\cite{DCNv2}, resembling a dynamic convolution rather than attention. 

\begin{table}[t]
\begin{center}
\caption{
Comparison on deformable attention in \ourmethod{} and in Deformable-DETR under different computational budgets. The GPU memory cost is measured in a forward pass with 64 batch size under inference mode.
}
\vspace{-0.5cm}
\label{tab:ddetr}
\newcommand{\tabincell}[2]{\begin{tabular}{@{}#1@{}}#2\end{tabular}}
\setlength{\tabcolsep}{1.1mm}{
\renewcommand\arraystretch{1.2}
\resizebox{\linewidth}{!}{
\begin{tabular}{cc|ccc|c}
\thickhline
\tabincell{c}{Deformable\\Attn. Types} & \#Keys & FLOPs & \#Param & \tabincell{c}{Peak\\Memory} & \tabincell{c}{IN-1K\\Top-1 Acc.} \\
\hline
\multirow{3}*{\tabincell{c}{Deformable\\DETR~\cite{DeformableDETR}}} & 16 & 4.17G & 22.1M & 11.5GB & 79.2 \\
& 49 & 4.84G & 25.6M & 20.1GB & 80.6 \\
& 100 & 5.89G & 31.2M & 33.6GB & 81.1 \\
\hline
\rowcolor{mygray}
\textbf{\ourmethod{}+} & 49 & 4.05G & 22.8M & 9.1GB & \textbf{83.0} \\
\thickhline
\end{tabular}}}
\vspace{-0.4cm}
\end{center}
\end{table}

\noindent
\textbf{Memory consumption.}
The deformable attention in Deformable DETR is not compatible with the dot product similarity computation for its huge memory footprints mentioned in Sec.~\ref{sec:m_def_attn}. Therefore, the linear predicted attention with linear space complexity is used to avoid computing dot products, along with a smaller number of keys $K=4$ to reduce the memory cost. 

\noindent
\textbf{Empirical results.}
To validate our claim, we replace the DMHA in \ourmethod{} with the modules in \cite{DeformableDETR} to show that a naive adaptation is inferior for the vision backbone. We use \ourmethod{}-T+ as a baseline, where the results are shown in Table~\ref{tab:ddetr}.
We vary the number of keys of the Deformable DETR model in 16, 49, and 100. The model $K\!=\!16$ has FLOPs and parameters to \ourmethod{}-T+, only achieving 79.2\% accuracy. When using the same number of keys in 49, the memory cost becomes over $2\!\times{}\!$ of \ourmethod{}+, but the performance does not catch up. As the number of keys increases to 100, the memory, FLOPs and parameters all grow drastically. However, the accuracy of the Deformable DETR attention still has a gap of 1.9\%.

\begin{figure*}
    \captionsetup[subfloat]{labelformat=empty,position=top}
    \hspace{0.1mm}
    \subfloat[Input Image]{\includegraphics[width=0.19\linewidth]{figures/motorbike_img.jpg}} \hspace{0.5mm}
    \subfloat[Instance Seg. Results]{\includegraphics[width=0.19\linewidth]{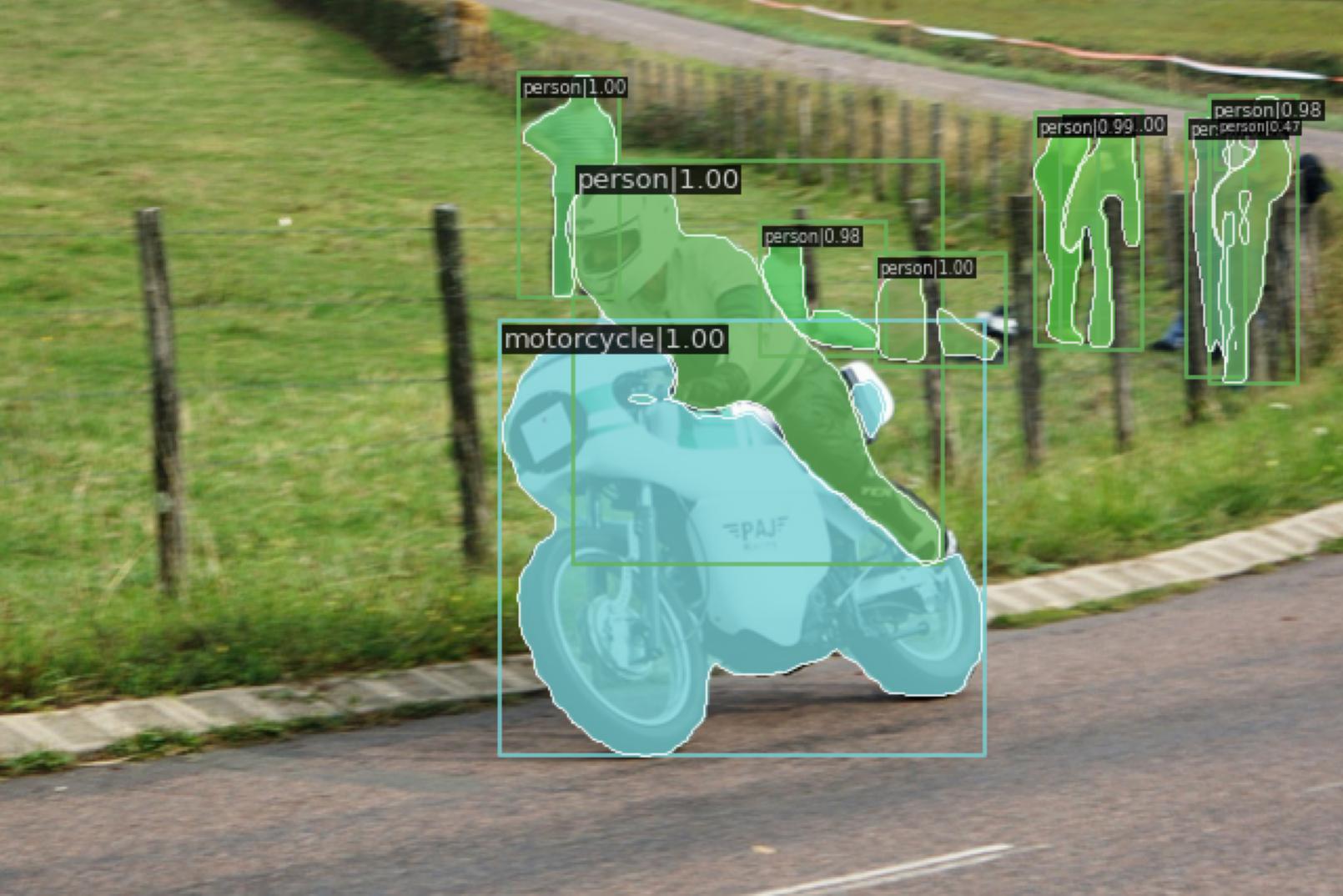}} \hspace{0.5mm}
    \subfloat[Overall Important Keys]{\includegraphics[width=0.19\linewidth]{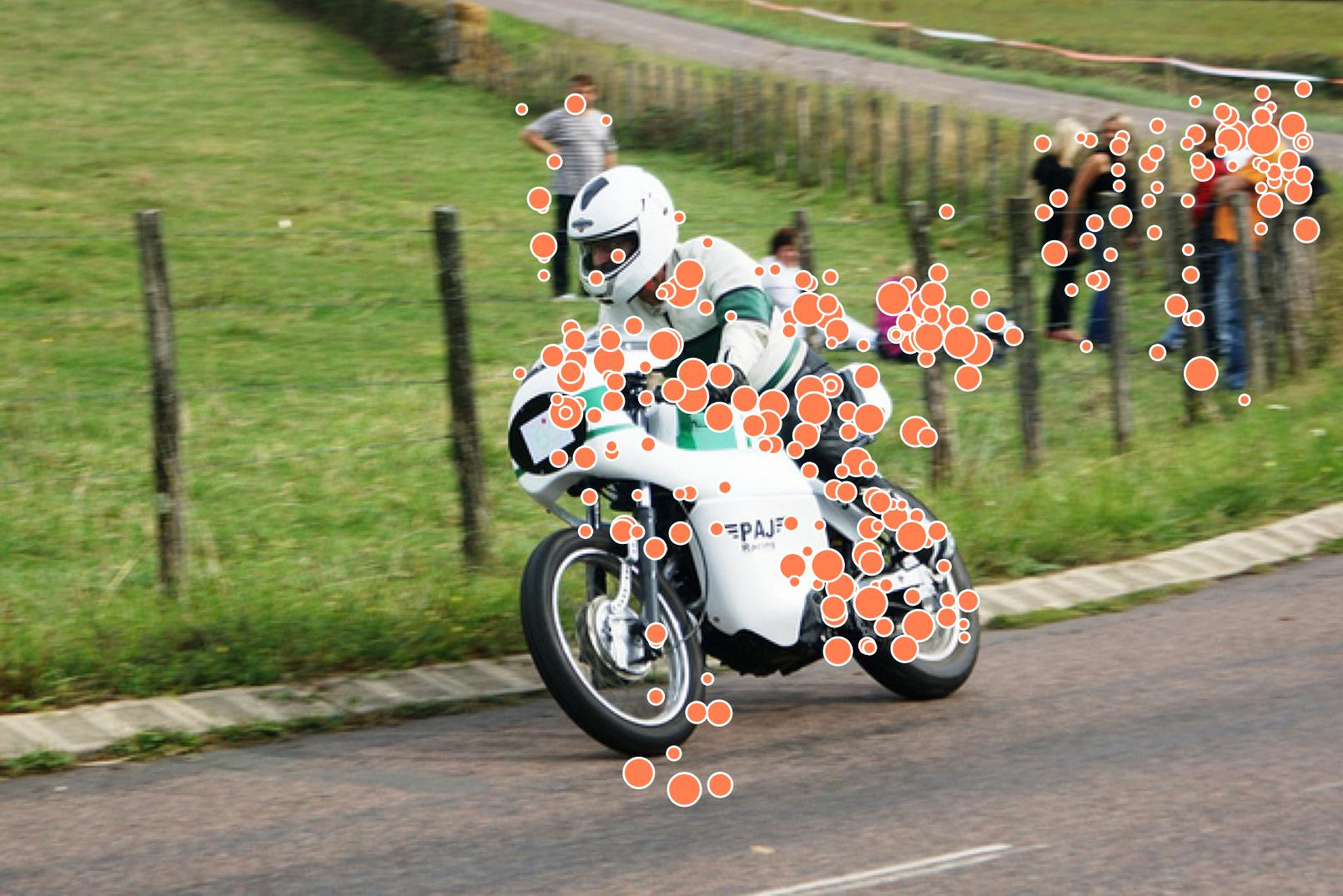}} \hspace{0.5mm}
    \subfloat[Important Keys to Query \#1]{\includegraphics[width=0.19\linewidth]{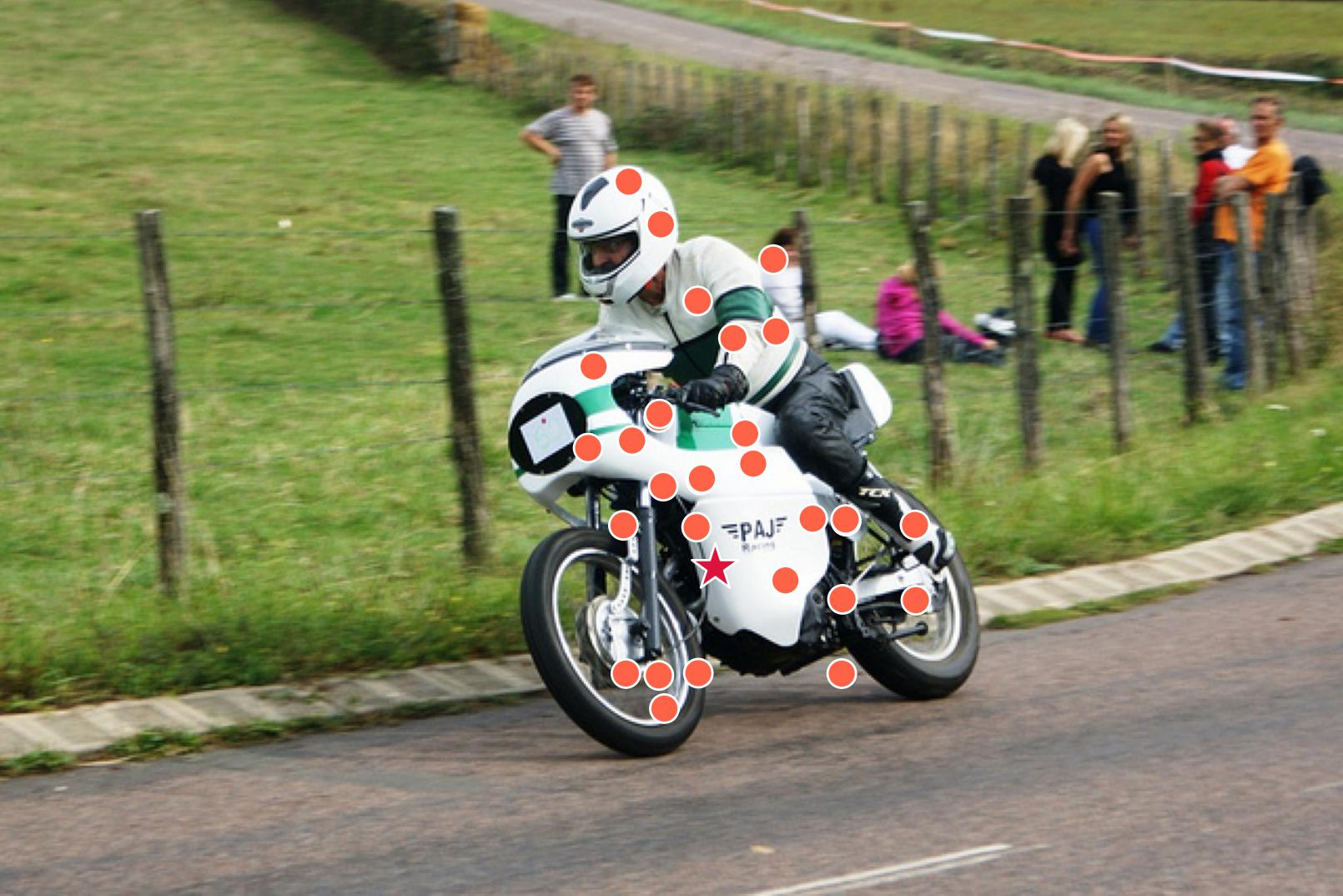}} \hspace{0.5mm}
    \subfloat[Important Keys to Query \#2]{\includegraphics[width=0.19\linewidth]{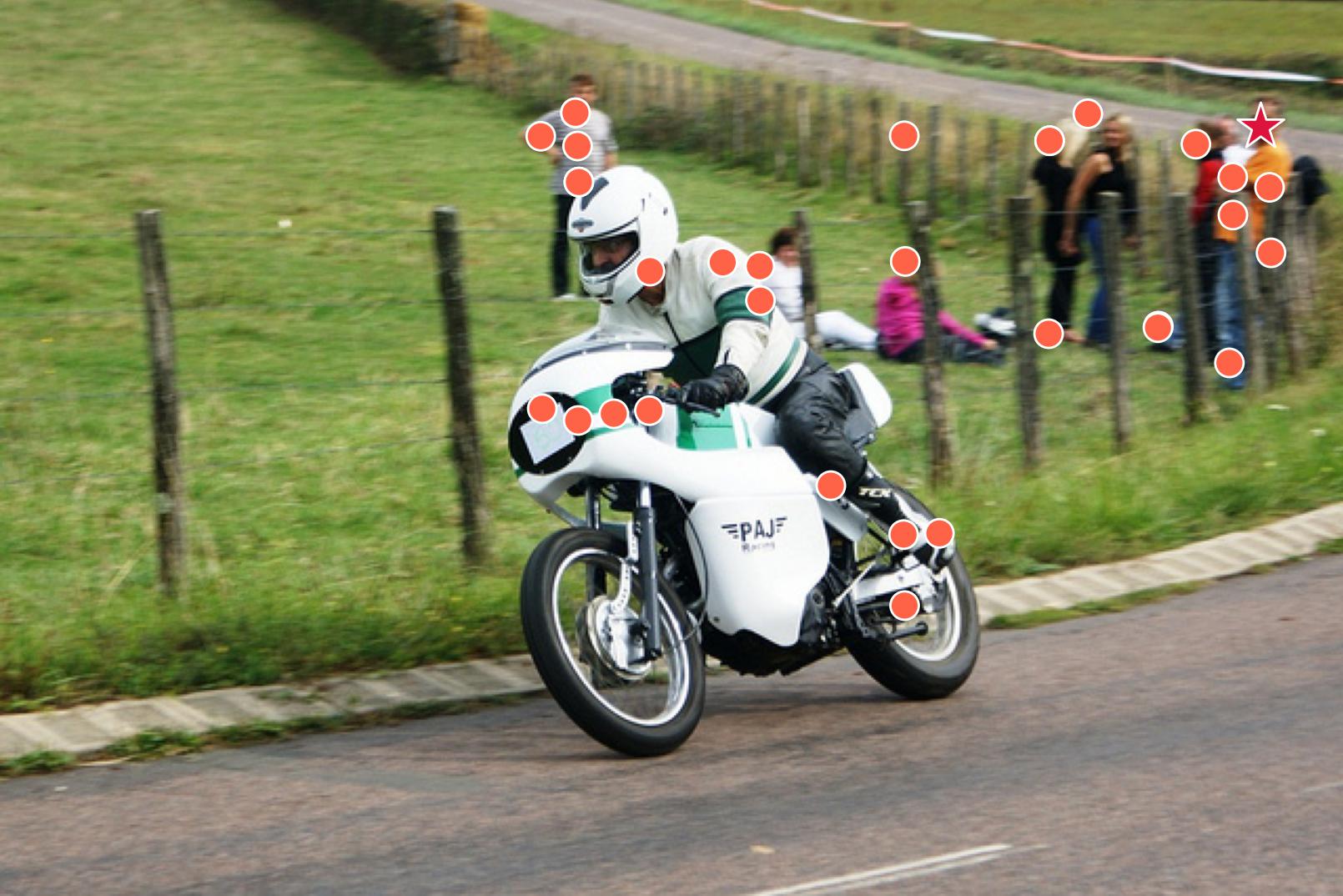}} \vspace{-0.1in} \\
    \vspace{-0.1in}
    \subfloat{\includegraphics[width=0.19\linewidth]{figures/elephant_img.jpg}} \hspace{0.5mm}
    \subfloat{\includegraphics[width=0.19\linewidth]{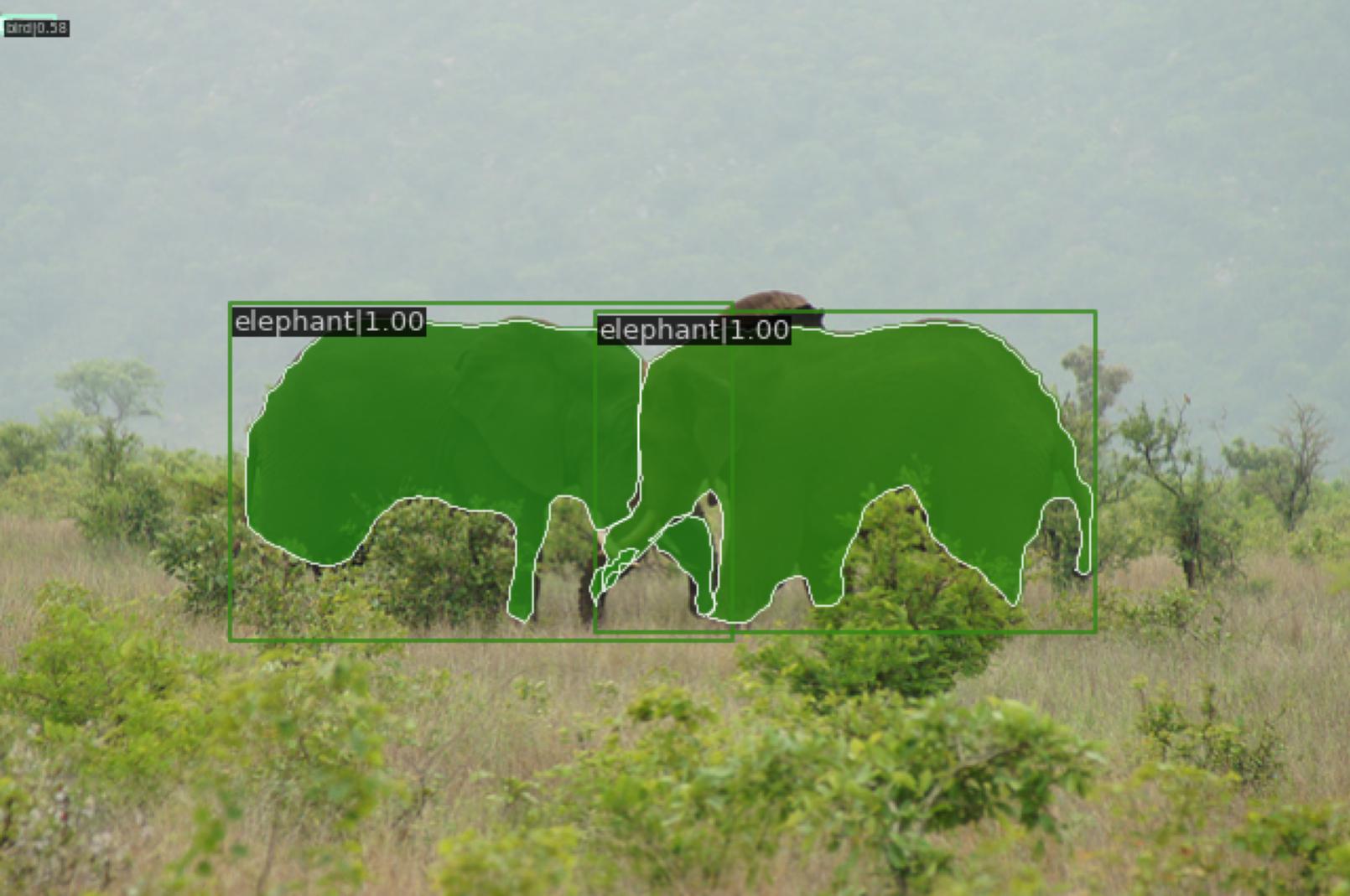}} \hspace{0.5mm}
    \subfloat{\includegraphics[width=0.19\linewidth]{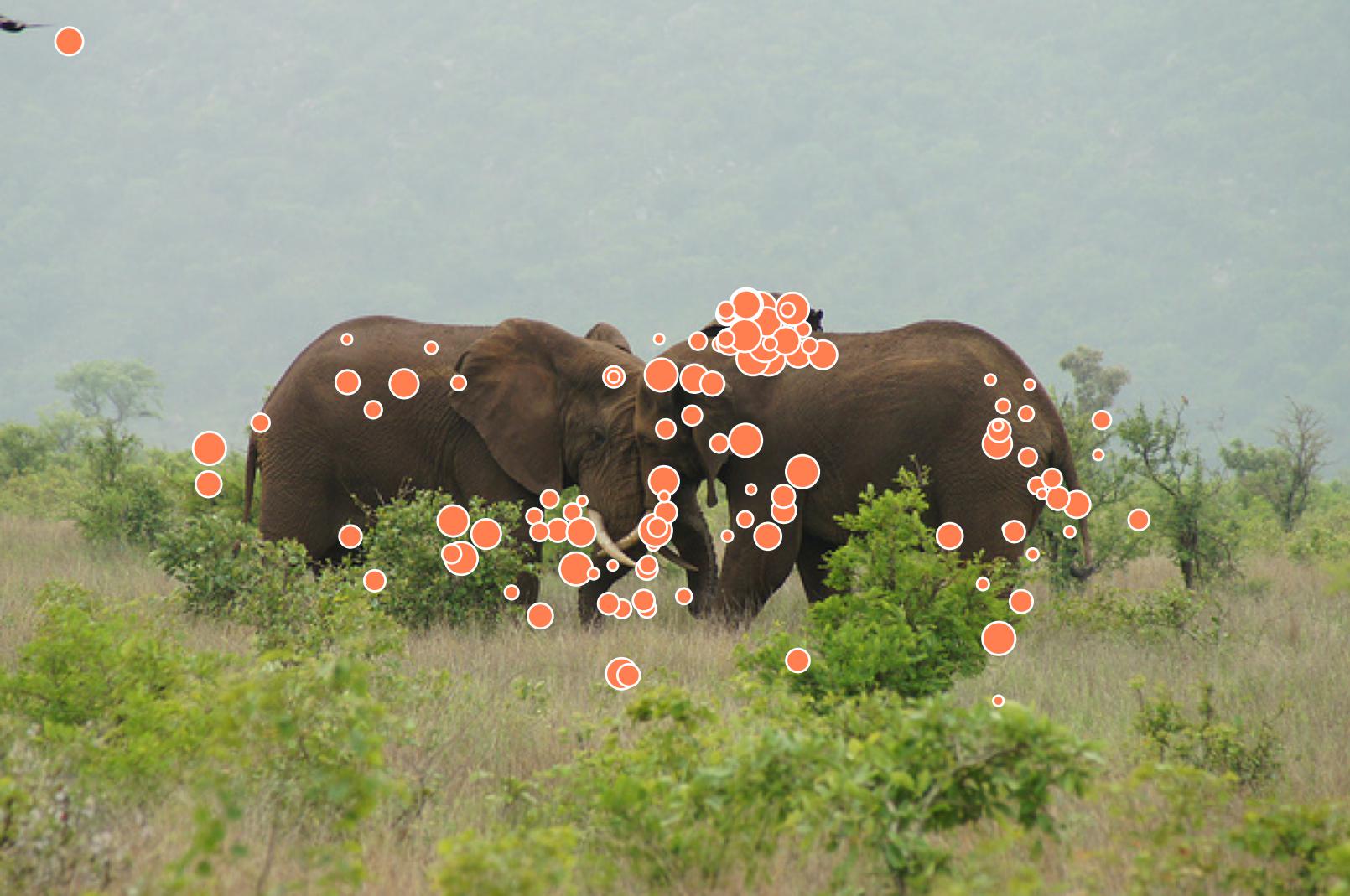}} \hspace{0.5mm}
    \subfloat{\includegraphics[width=0.19\linewidth]{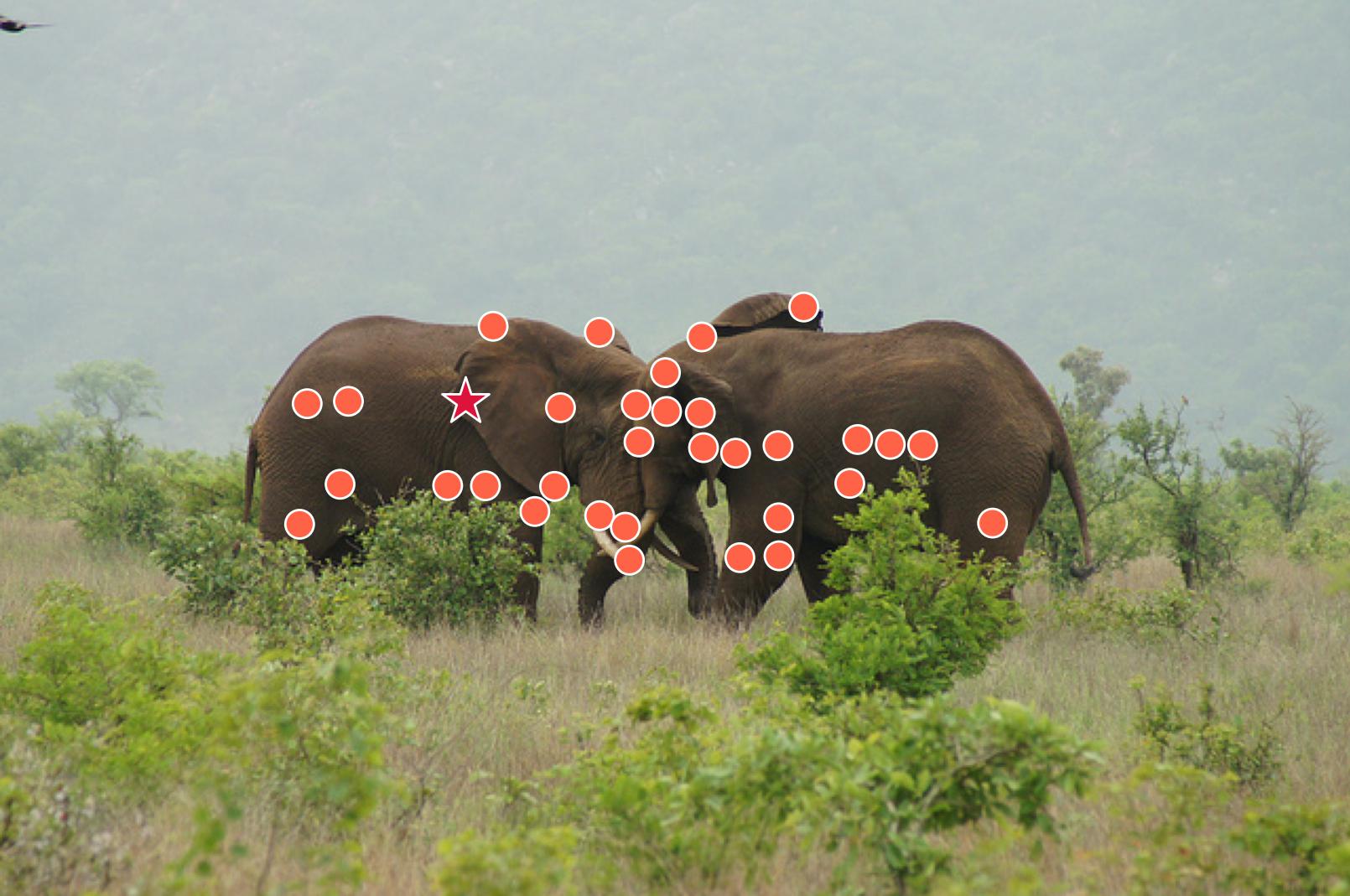}} \hspace{0.5mm}
    \subfloat{\includegraphics[width=0.19\linewidth]{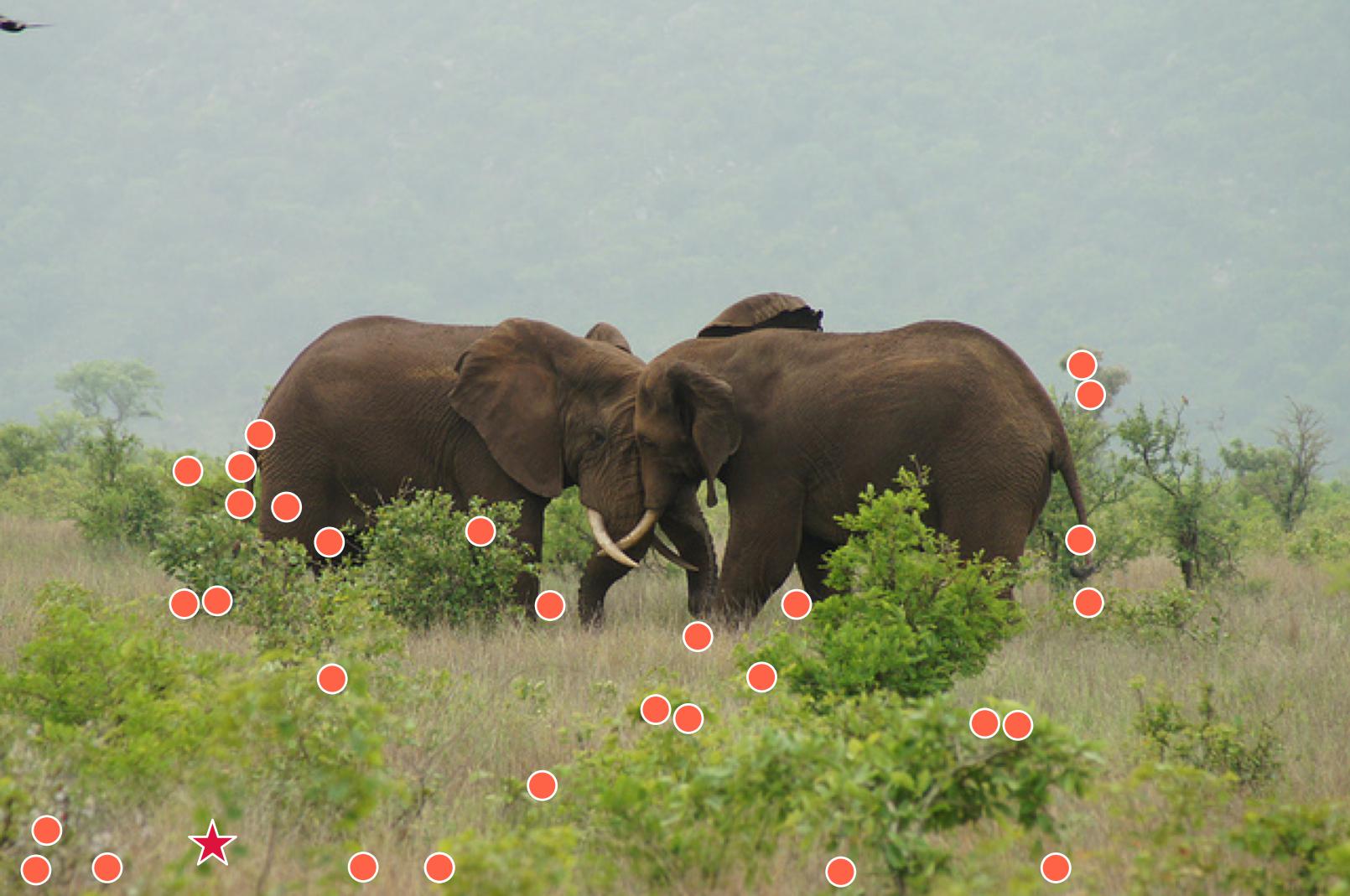}} \\
    \vspace{-0.1in}
    \subfloat{\includegraphics[width=0.19\linewidth]{figures/baseball_img.jpg}} \hspace{0.5mm}
    \subfloat{\includegraphics[width=0.19\linewidth]{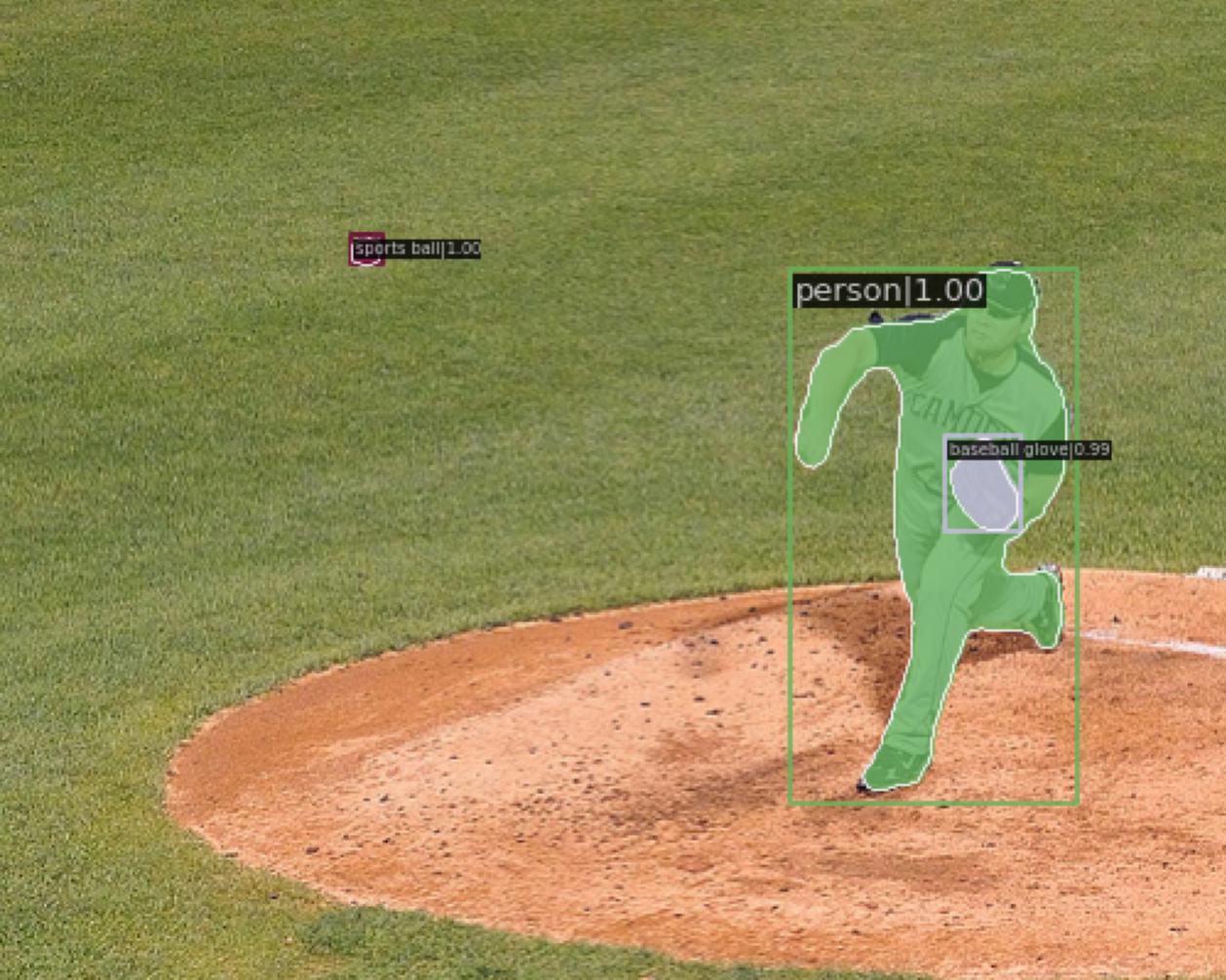}} \hspace{0.5mm}
    \subfloat{\includegraphics[width=0.19\linewidth]{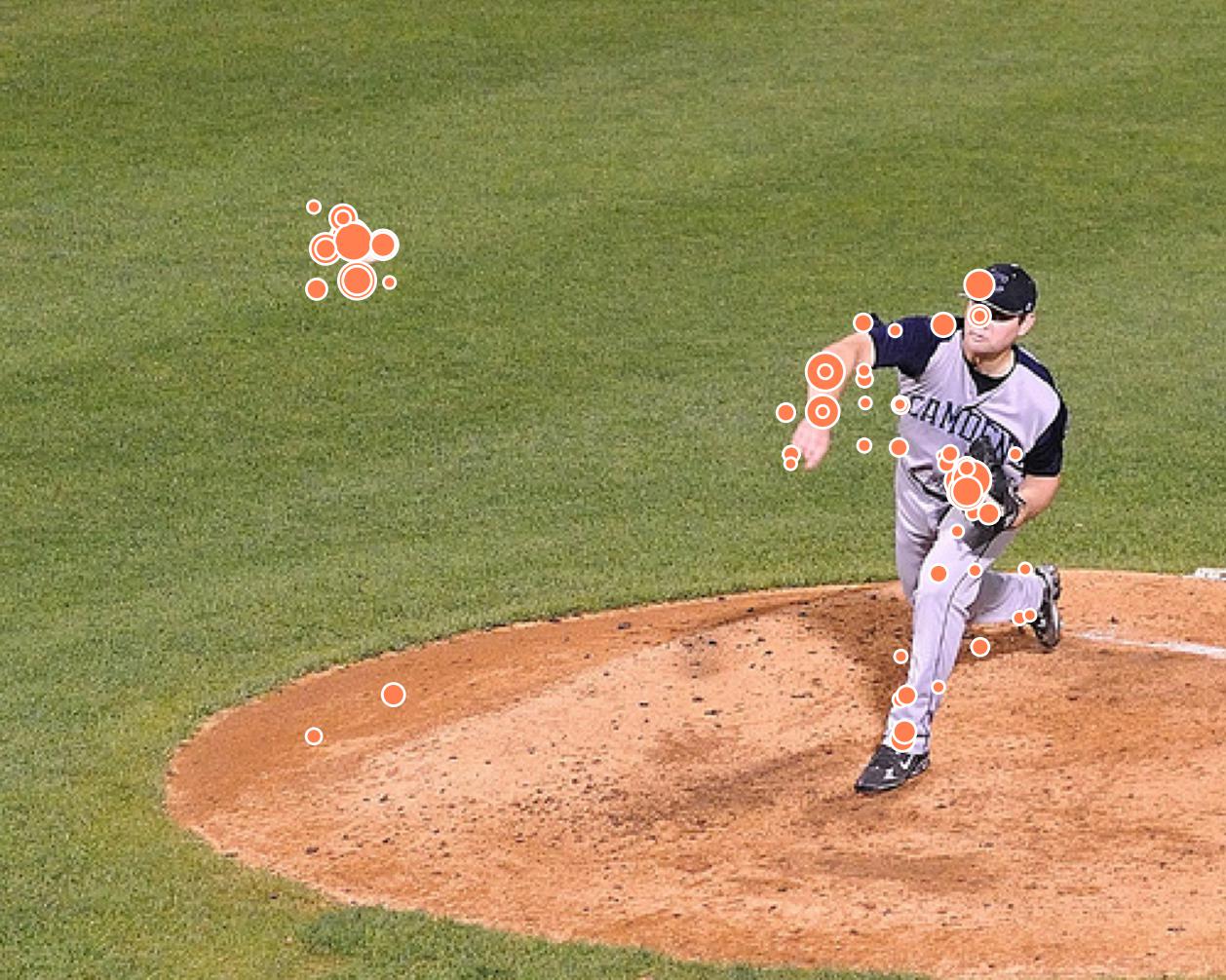}} \hspace{0.5mm}
    \subfloat{\includegraphics[width=0.19\linewidth]{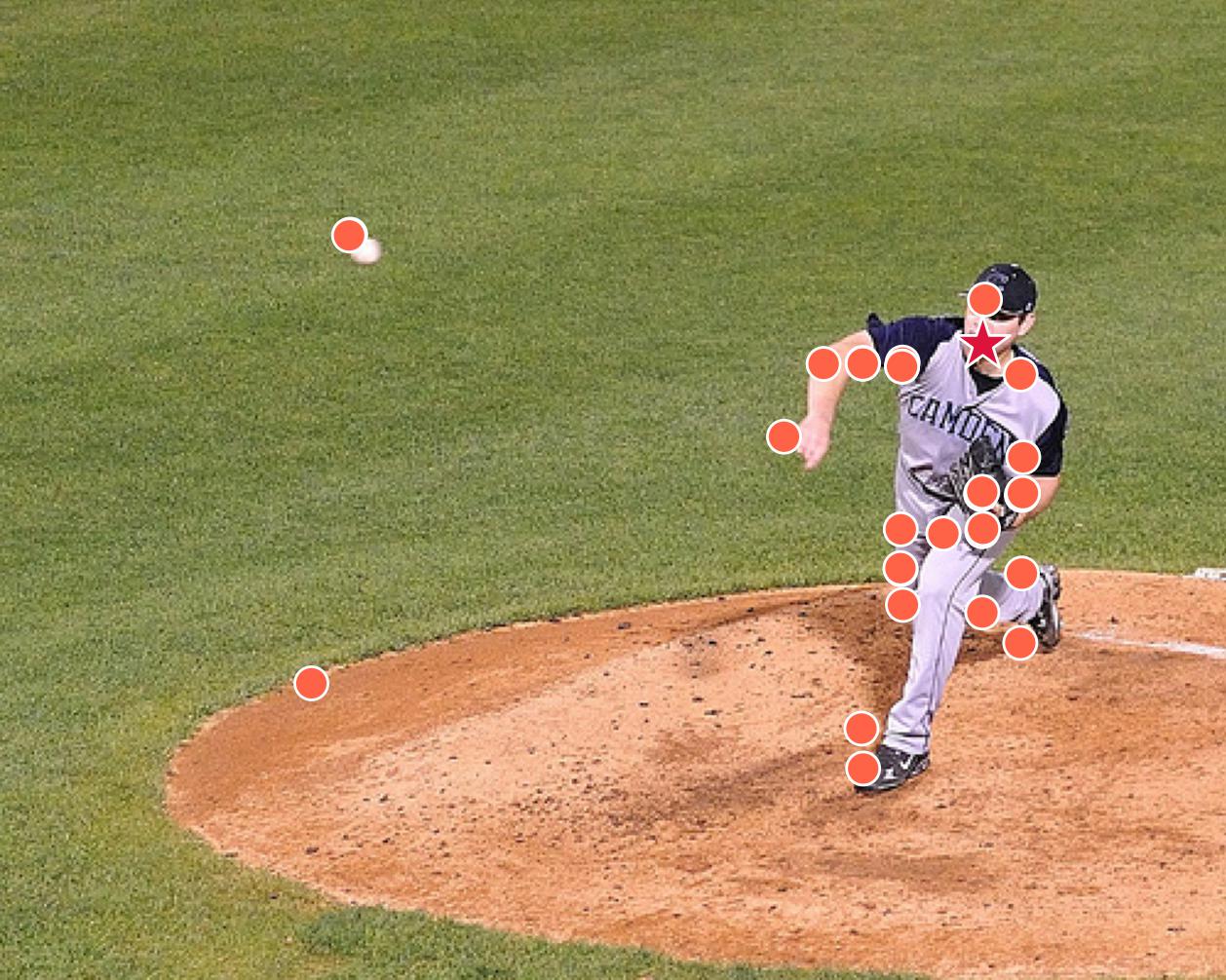}} \hspace{0.5mm}
    \subfloat{\includegraphics[width=0.19\linewidth]{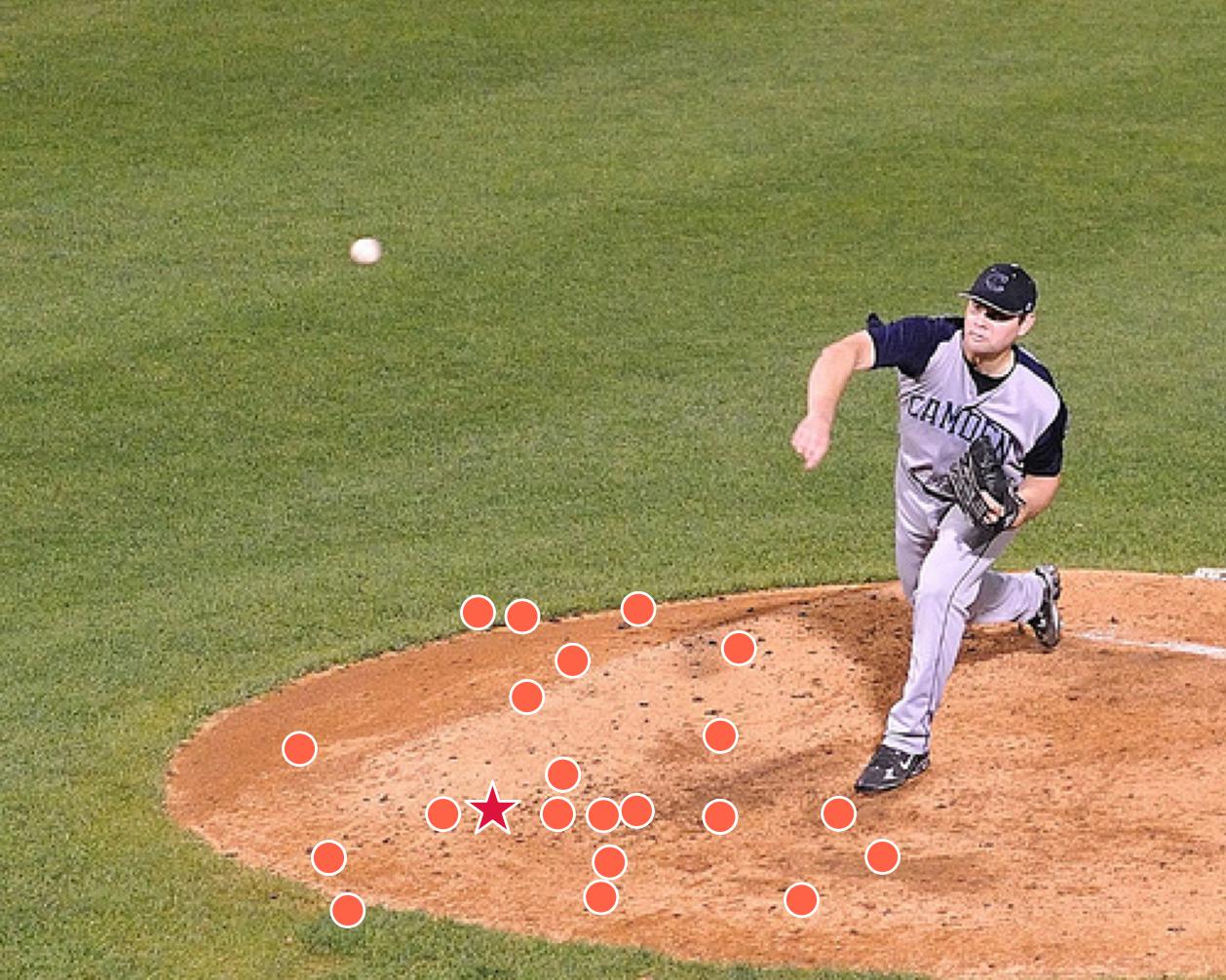}} \\
    \vspace{-0.1in}
    \subfloat{\includegraphics[width=0.19\linewidth]{figures/skiing_img.jpg}} \hspace{0.5mm}
    \subfloat{\includegraphics[width=0.19\linewidth]{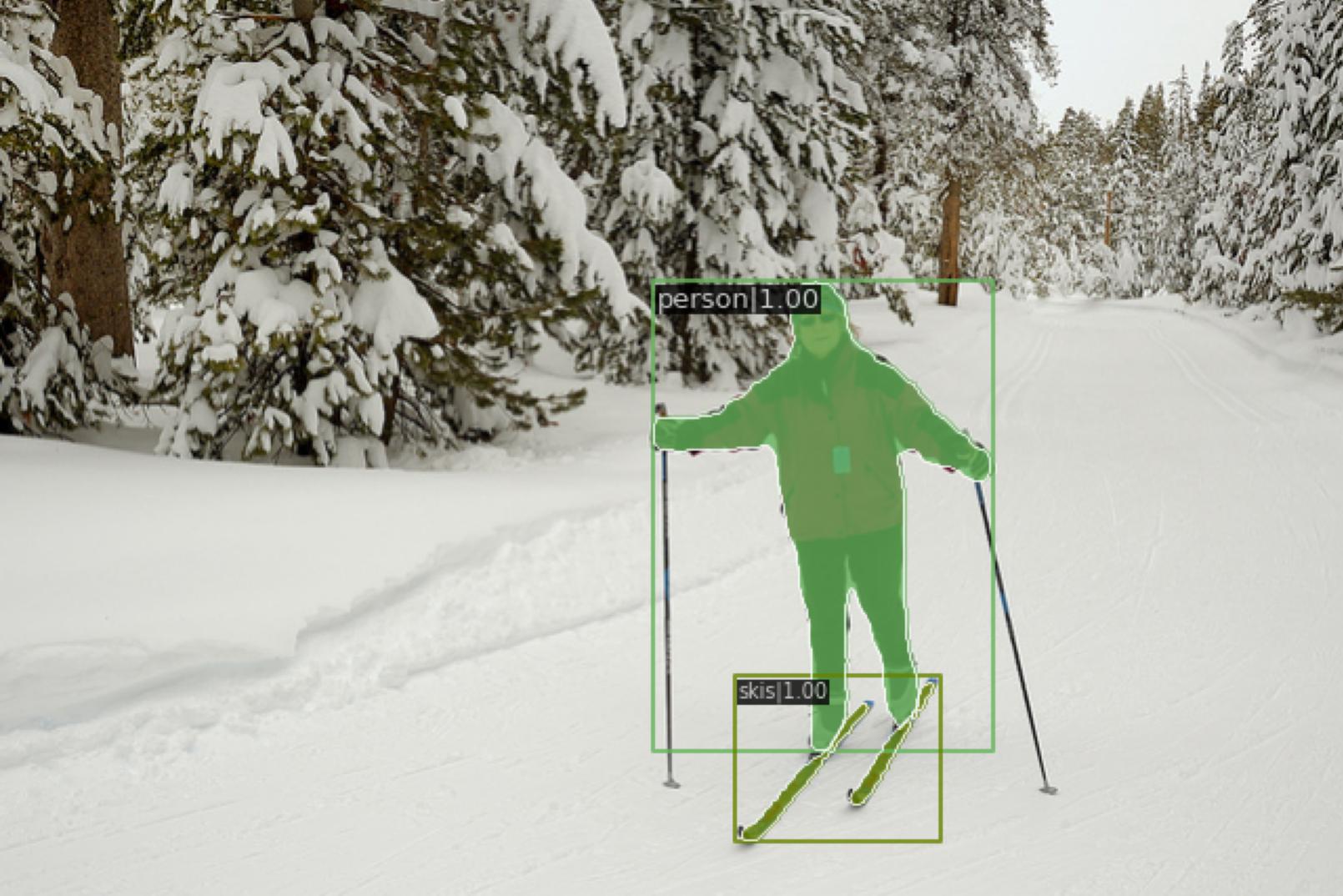}} \hspace{0.5mm}
    \subfloat{\includegraphics[width=0.19\linewidth]{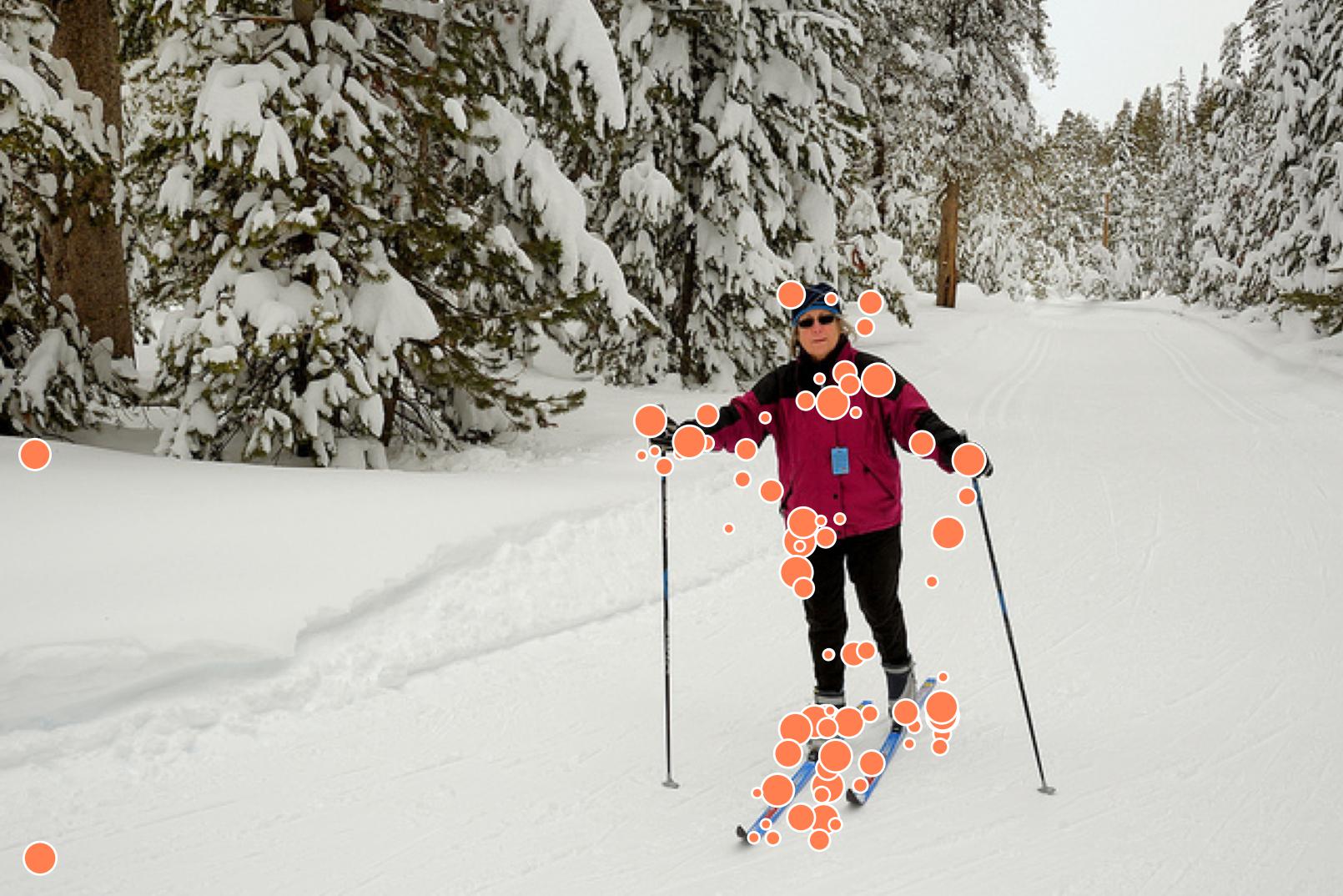}} \hspace{0.5mm}
    \subfloat{\includegraphics[width=0.19\linewidth]{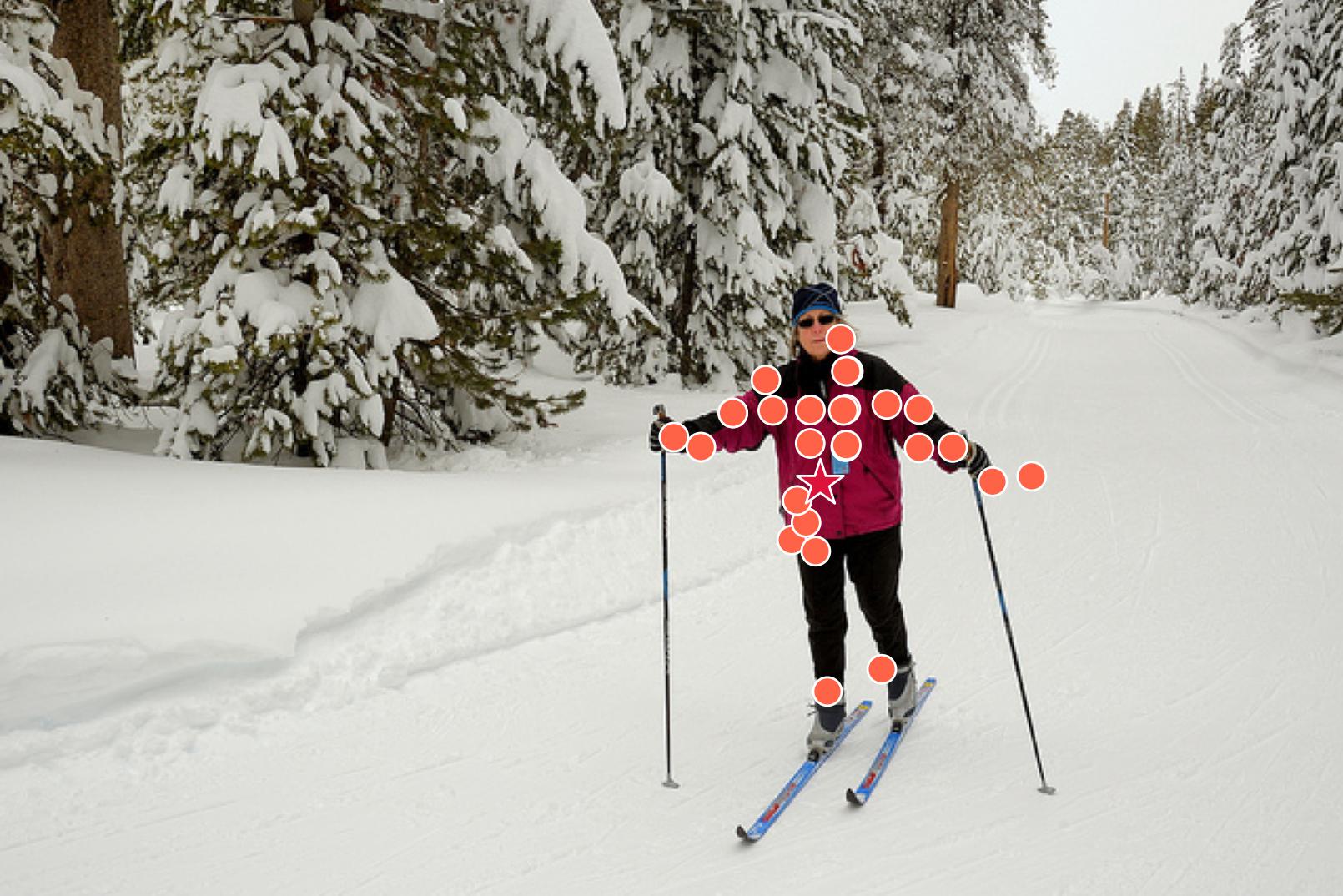}} \hspace{0.5mm}
    \subfloat{\includegraphics[width=0.19\linewidth]{figures/skiing_q02.jpg}} \\
    \vspace{-0.1in}
    \subfloat{\includegraphics[width=0.19\linewidth]{figures/skateboarding_img.jpg}} \hspace{0.5mm}
    \subfloat{\includegraphics[width=0.19\linewidth]{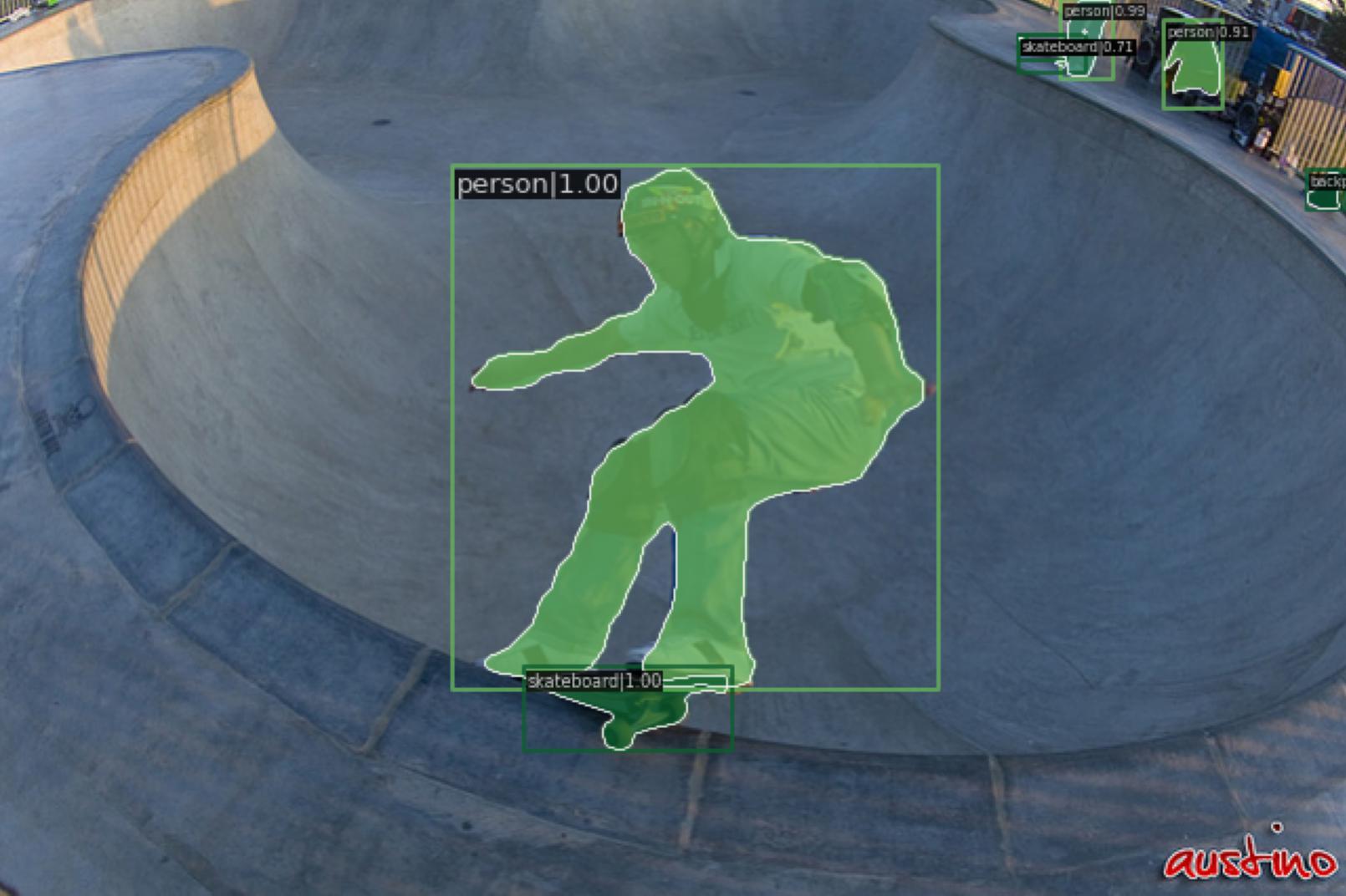}} \hspace{0.5mm}
    \subfloat{\includegraphics[width=0.19\linewidth]{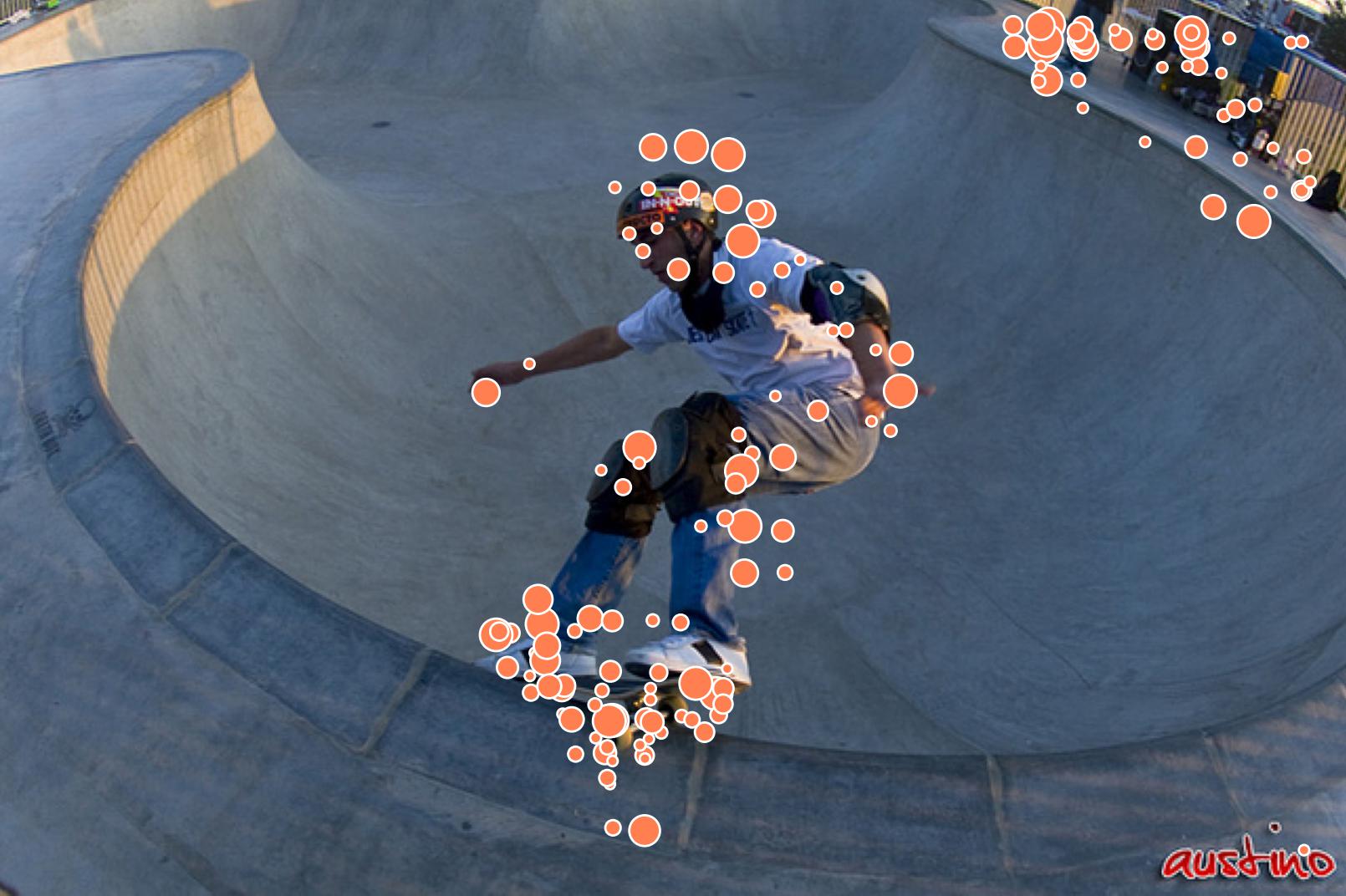}} \hspace{0.5mm}
    \subfloat{\includegraphics[width=0.19\linewidth]{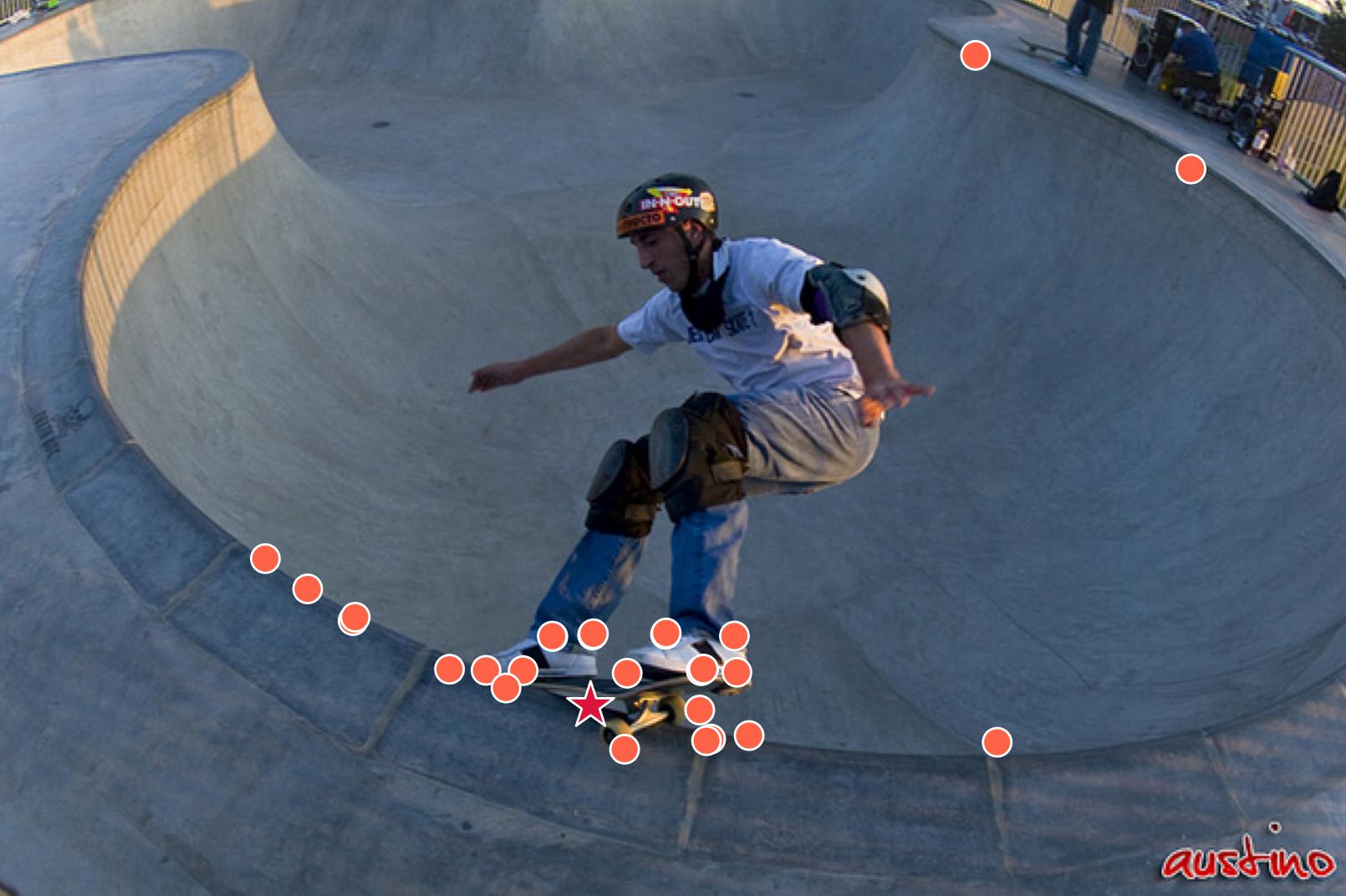}} \hspace{0.5mm}
    \subfloat{\includegraphics[width=0.19\linewidth]{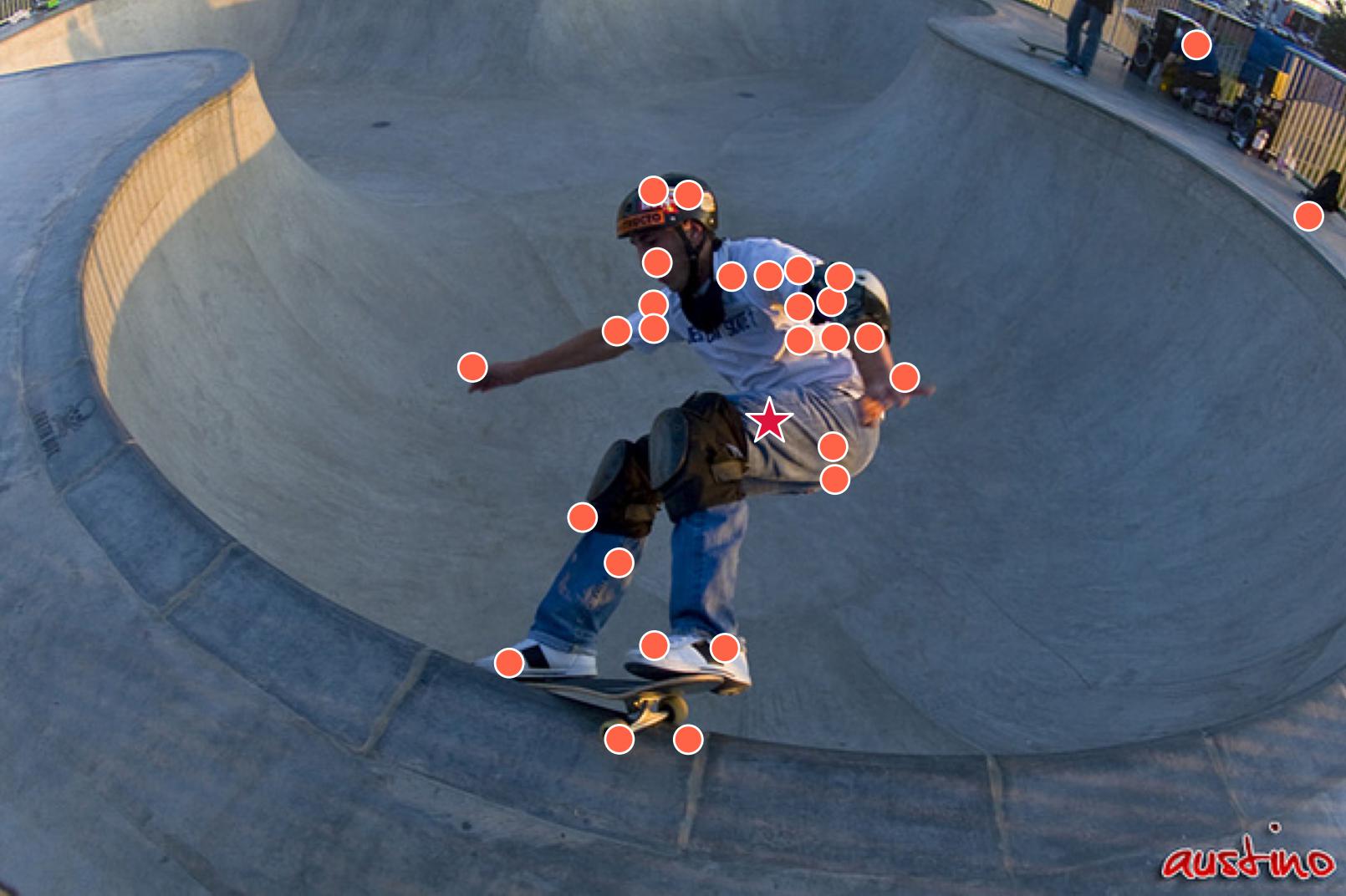}} \\
    \caption{Visualization of \ourmethod{}++ with Mask R-CNN~\cite{mrcn} on MS-COCO~\cite{coco} validation set. From left to right, each column displays the input image, the bounding boxes and masks of instance segmentation, the overall important keys of each attention head, and the important keys to two given query points, respectively. In the 3$^\text{rd}$ column, each orange circle denotes a deformed key, whose larger radius reflects higher attention score. In the 4$^\text{th}$ and 5$^\text{th}$ columns, the query point is marked as a red star, and the keys with high attention scores are marked in coral. \textbf{Zoom in for best view.}}
	\label{fig:vis}
    \vspace{-0.3cm}
\end{figure*}

\subsection{Visualization}\label{sec:vis}

We visualize some qualitative results of \ourmethod{}++ to verify the effectiveness of the proposed deformable attention, as shown in Figure~\ref{fig:vis}. We present five examples from MS-COCO~\cite{coco} validation dataset, using Mask R-CNN~\cite{mrcn} (3x) equipped with \ourmethod{}-T++ to perform instance segmentation. The first columns show the input images and the results output from the model. 

The 3$^\text{rd}$ column of Figure~\ref{fig:vis} displays the important keys in the image, whose scores are represented by the sizes of the orange circles, where a larger circle indicates a higher score. For a comprehensive analysis of the importance score, we cumulatively multiply the attention weights of the deformed keys from the last DMHA layer to previous layers and then average them among all queries to discover the keys with the most contributions. We observe that the proposed deformable attention learns to attend the important keys mainly in the foreground, \textit{e.g.}, in the 1$^\text{st}$ row, the keys with Top-10 accumulated attention scores are mostly distributed in the foreground motorbike and persons. Similar patterns are also found in the scene of the elephants, the baseball and pitcher, the skier and skis, and the skater and boards in the 2$^\text{nd}$ to 5$^\text{th}$ rows, indicating that the deformable attention can focus on the important regions of the objects, which supports our hypothesis illustrated in Figure~\ref{fig:fig1}.

In addition to the overall distribution of the deformed keys, we also provide visualization of the attention maps \textit{w.r.t.} some specific query tokens, depicted in the last two columns of Figure~\ref{fig:vis}. Taking the ski image in the 4$^\text{th}$ as an example, we place the query \#1 (red star) on the skier, and the keys with high attention scores (coral circles) are concentrated mainly on the body of the skier. In comparison, for query \#2 in the pine tree surround shown in the 5$^\text{th}$ column, its corresponding deformed keys scatter among the background trees, further demonstrating that \ourmethod{}++ focuses on the keys that are shifted to the relevant parts of the queries with meaningful offsets.
\section{Conclusion}\label{sec:conclusion}

This paper presents \omlong{}, a novel hierarchical Vision Transformer that can be adapted to various visual recognition tasks, including image classification, object detection, instance segmentation, and semantic segmentation. With the proposed deformable attention and DMHA module, \ourmethod{} and \ourmethod{}++ are capable of learning spatially dynamic attention patterns in a data-dependent way and modeling geometric transformations. Extensive experiments demonstrate the effectiveness of \ourmethod{}++ over other competitive ViT/CNNs. We hope our work can inspire insights towards designing dynamic visual attention mechanisms and more powerful foundation models. 

\ifCLASSOPTIONcompsoc
  \section*{Acknowledgments}
\else
  \section*{Acknowledgment}
\fi

This work is supported in part by the National Key R\&D Program of China under Grant 2021ZD0140407, in part by the National Natural Science Foundation of China under Grants 62022048 and 62276150, and in part by the Guoqiang Institute of Tsinghua University. We also appreciate the donation of computational resources from Hangzhou High-Flyer AI Fundamental Research Co.,Ltd.


{
\bibliographystyle{IEEEtran}
\bibliography{egbib}
}





\end{document}